\documentclass{article}

\usepackage{arxiv}

\usepackage[utf8]{inputenc} 
\usepackage[T1]{fontenc}    
\usepackage{hyperref}       
\usepackage{url}            
\usepackage{booktabs}       
\usepackage{amsfonts}       
\usepackage{nicefrac}       
\usepackage{microtype}      
\usepackage{lipsum}		
\usepackage{graphicx}
\usepackage{natbib}
\usepackage{doi}
\usepackage{subfig}

\title{Evaluating Deep Neural Networks for Image Document Enhancement}

\author{%
  \href{https://orcid.org/0000-0001-6611-5327}{\includegraphics[scale=0.06]{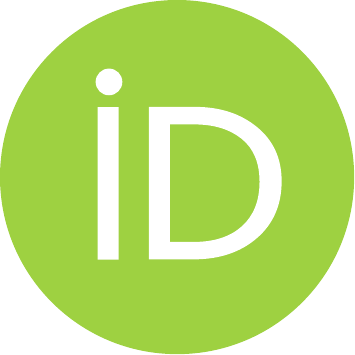}\hspace{1mm}Lucas N.~Kirsten} \qquad
  \href{https://orcid.org/0000-0003-0694-0311}{\includegraphics[scale=0.06]{orcid.pdf}\hspace{1mm}Ricardo Piccoli} \qquad
  \href{https://orcid.org/0000-0002-9579-1663}{\includegraphics[scale=0.06]{orcid.pdf}\hspace{1mm}Ricardo Ribani}
  \\ \\
  Department of Print Software\\
  HP Inc. -- R\&D\\
  Porto Alegre -- RS, 90619-900, Brazil \\
  \footnotesize
  \texttt{\{lucas.nedel.kirsten, ricardo.piccoli, ricardo.ribani\}@hp.br}
}

\bgroup\catcode`\#=12\egroup \renewcommand{\shorttitle}{Evaluating Deep Neural Networks for Image Document Enhancement}

\hypersetup{
pdftitle={Evaluating Deep Neural Networks for Image Document Enhancement},
pdfauthor={Lucas N.~Kirsten~et.al.},
pdfkeywords={document analysis, image enhancement, deep learning},
}

\begin{document}
\maketitle

\begin{abstract}
	This work evaluates six state-of-the-art deep neural network (DNN) architectures applied to the problem of enhancing camera-captured document images. The results from each network were evaluated both qualitatively and quantitatively using Image Quality Assessment (IQA) metrics, and also compared with an existing approach based on traditional computer vision techniques. The best performing architectures generally produced good enhancement compared to the existing algorithm, showing that it is possible to use DNNs for document image enhancement. Furthermore, the best performing architectures could work as a baseline for future investigations on document enhancement using deep learning techniques. The main contributions of this paper are: a baseline of deep learning techniques that can be further improved to provide better results, and a evaluation methodology using IQA metrics for quantitatively comparing the produced images from the neural networks to a ground truth.
\end{abstract}


\section{Introduction}
Recent advances in deep learning architectures have given rise to novel image editing applications, such as image-to-image translation. Essentially, an image-to-image translation network can be trained to perform an arbitrary transformation on an input image, which is mainly determined by the training data~\citep{image_translation}. This has led to the use of deep learning in many computer vision and image processing applications, such as image re-targeting, enhancement, restoration and image generation~\citep{img_denoiser, image_translation}. The main advantage of using these techniques is their ability to learn complex patterns from the training data, and generalizing for previously unseen data.

In this paper, we focus on the specific task of document image enhancement, \emph{i.~e.}, performing automatic corrections and enhancing the digital version of a document image that has been captured with a camera, focusing on improve their perceived visual quality. This can be achieved by eliminating noise, sharpening the image, enhancing contrast, among other techniques~\citep{image_enhancement}. There are many factors that affect image quality when digitizing a document, such as lighting conditions in the environment during capture, and resolution. These factors play a crucial role when producing a physical copy of a digital document, because it directly affects the quality of the output medium, and thus needs to be optimized for a particular case (\emph{e.~g.}, ink on paper). Therefore, specialized hardware such as flatbed scanners are commonly used to account for the variability of these factors and minimize their impact on the final image.

Herein, we study existing deep learning models adapted to work in the context of \emph{camera-captured} documents image enhancement, focusing on the following basic aesthetic corrections: background lighting removal, content color preserving, tone mapping and contrast enhancement. We assume that dedicated scanning hardware is not readily available, and a consumer/smartphone camera has been used for image acquisition. This is usually the case for mobile scanning applications, where users may digitize a document by taking a picture of a document using a mobile or smartphone camera.

The goal of this work is thus to investigate deep learning techniques that can be adapted to work on the document enhancement task. More into it, we adapted and trained the following state-of-the-art neural networks: U-Net~\citep{unet}, EDSR~\citep{edsr}, WDSR~\citep{wdsr}, SRGAN~\citep{srgan}, IRCNN~\citep{ircnn} and the CycleGAN~\citep{cyclegan}, and compared their outputs with the approach proposed by \cite{pagelift} to enhance scanned documents using traditional computer vision techniques, using the following metrics: PSNR, MS-SSIM~\citep{ssim}, PIE~\citep{pie} and WaDIQaM~\citep{wadiqam}. Our main contributions to the field are: a baseline of DL techniques that can be further improved to provide better results, and an evaluation methodology using Image Quality Assessment (IQA) metrics~\citep{iqa, iqa_survey} for quantitatively comparing the produced images from the neural networks to a ground truth.

The rest of this paper is organized as follows: Section~\ref{sec:related-work} briefly discusses related work on the topics of image enhancement and image quality assessment. Section~\ref{sec:experimental-setup} describes the design of the ground truth and experiments in detail, including details on how each different model was trained and fine-tuned. Finally, Sections~\ref{sec:results} and~\ref{sec:conclusion} present the results obtained for each model and quality metric tested and our final considerations regarding these results.

\section{Related work}\label{sec:related-work}
\subsection{Image enhancement}
Recent convolutional neural networks (CNN) based methods have been demonstrated to be conductive for image enhancement~\citep{img_enh1, img_enh2, img_enh3}, showing that a variety of different techniques can be used for such task. For instance, enhancement may be encoded as identifying a noise pattern that should be removed from the input image, which in turn can be solved with existing image denoising techniques~\citep{img_denoiser}.

Another family of techniques to improve image quality are super resolution-based approaches, where the goal is to improve the quality of an up-scaled version of the input image~\citep{superresolution_survey}. While classic techniques (such as Bi-linear filtering or Nearest-neighbor interpolation) use local interpolation methods to fill the gaps created when increasing the image size, super-resolution solutions aim to fill these gaps with more suitable values by learning the patterns of the image globally and, thus preserving aspects such as the image gradient. This is usually achieved by using robust feature extractor modules for a particular neural network architecture~\citep{edsr, wdsr, srgan}.

For the document enhancement task in particular, \cite{pagelift} and \cite{dropbox} have addressed this problem by employing a traditional computer vision workflow to enhance the document content and correct overall lighting differences present in the document image. Conversely, \cite{deepotsu} and \cite{doc_enhancement} used neural networks with specialized architectures to address the problem.

\subsection{Image Quality Assessment (IQA)}

Existing approaches for measuring image quality can be split into two basic categories: reference, and reference-less or blind~\citep{iqa,iqa_survey}. Reference methods use a ground-truth (\emph{i.~e.}, a clean image) and its distorted version (\emph{i.~e.}, with undesirable artifacts added to the image) to compute a score, while reference-less ones use only the distorted image. For the purpose of evaluating techniques for document enhancement, only reference methods are considered herein. In this manner, it is possible to compare  enhanced document images to their native format (\emph{i.~e.}, a digital version in the PDF format).

Reference-based IQA methods range from statistical measures commonly applied for comparing pairs of images, such as the Mean Absolute Error and the Mean Squared Error, which are pixel-wise metrics (\emph{i.~e.}, the final value is derived by comparing each pair of corresponding pixels between the images), to methods based on structural analysis, such as the Multi-Scale Structural Similarity (MS-SSIM)~\citep{ssim}, and even approaches based on modern deep learning techniques, such as WaDIQaM~\citep{wadiqam} and Perceptual Image Error (PIE)~\citep{pie}. All of these techniques attempt to return a score based on human perception of image quality~\citep{iqa_hvs}. However, human perception can vary drastically from person to person, depending on the application and intention of the image usage. For this reason, each IQA metric tends to work best for different sets of images and distortions.

For document image quality assessment (DIQA), \cite{doc_iqa} and \cite{doc_iqa2} use Optical Character Recognition (OCR) engines for verifying aspects such as text alignment, text color and size, that are not relevant for the context of this work. \cite{diqa1}, on the other hand, proposed a metric that considers aesthetic aspects of the document image, however, we have not considered it since it is a reference-less metric. \cite{diqa2} proposed a metric based on measuring human perception over quality and document readability, but we could not find an implementation available in order to reproduce the results of that paper.

\section{Experimental setup}\label{sec:experimental-setup}

\subsection{Datasets}

For proper training and validation of each network, we used two distinct sets of images. We first built a small dataset by collecting 198 samples of digital documents in the PDF format. Each document was then printed and subsequently captured using smartphone cameras, pairing each captured image with its native PDF version.

At the time, we did not have any equipment readily available to automate the image capture process, and thus we opted to use a private dataset of scanned documents. This dataset is composed of several types of documents such as plain text, magazines, articles, advertising documents, flyers; in black and white or including colors; and also with variations of shadows. In order to acquire the enhanced version of these images (\emph{e.g.} the ground truth), we used an existing improved implementation of the algorithm proposed by \cite{pagelift}. We then manually inspected each output in order to discard poorly enhanced images. A total of 3000 image pairs were produced using this method.

Finally, we used the first set of 198 images (comprising pairs of photographed and PDF images) as the test set, and the larger set of 3000 images (comprising pairs of raw images and their enhanced versions with the \cite{pagelift} algorithm) as the training set.

\begin{figure}[h]
    \footnotesize
    \centering
	\includegraphics[width=0.8\textwidth]{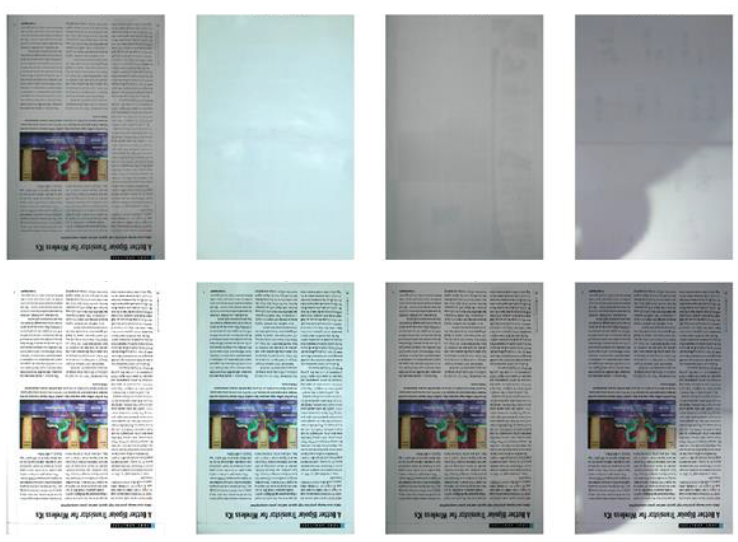}
	\caption{Lighting surface augmentation method. First column: original raw image and enhanced image with algorithm from \cite{pagelift}, respectively. Top: lighting surfaces extracted from other images; Remaining images at the Bottom: resulting augmented raw images using the upper lighting surfaces and the enhanced image.}
	\label{fig:lighting_surface_aug}
\end{figure}

Since our training dataset has few samples for training neural networks algorithms (that usually require a huge amount of data), we employed two augmentation techniques. The first one synthetically added illumination surfaces to each of the enhanced documents. The surface images were obtained by leveraging the intermediate step of lighting correction of the \cite{pagelift} algorithm and saving the illumination (or \emph{lighting}) surface as separate images. To apply these surfaces to a clean image, we simply multiply the values of the two images pixel-wisely, following a Retinex model of color constancy~\citep{pagelift}. If $\mathrm{E}\in\mathbb{R}^3$ is an enhanced image with lighting-corrected content, then we have:
\begin{equation}
	\mathrm{E}(x,y,c) = \mathfrak{F}\left( \frac{\mathrm{R}(x,y,c)}{\mathrm{L}(x,y,c)} \right) ,
\end{equation}
where $\mathrm{R}\in\mathbb{R}^3$ is the raw (scanned) image, $\mathrm{L}\in\mathbb{R}^3$ is the illumination surface of the raw image, and $\mathfrak{F}:\mathbb{R}^3\rightarrow\mathbb{R}^3$ is an enhancement function. Assuming that $\mathfrak{F}$ is a linear mapping function, we can obtain an image with added illumination by:
\begin{equation} \centering
	\mathrm{R}(x,y,c) = \mathrm{E}(x,y,c) \cdot \mathrm{L}(x,y,c) .
\end{equation}
Figure~\ref{fig:lighting_surface_aug} shows an example of this process. Using this augmentation strategy, it is easy to see that, for each image in our training set, we are able to produce $3000^{3000} \approx 10^{10432}$ variations of it.

The second employed augmentation technique produced randomly cropped pair versions of the raw and enhanced images. However, as is the case for several documents, most crops do not contain any text/graphical content (only background). This is a problem, as it may produce a biased network. To remove these crops, we used image gradients computed with a Laplacian kernel\footnote{We used the following OpenCV implementation of the Laplacian filter: \url{https://docs.opencv.org/3.4/d4/d86/group__imgproc__filter.html\#gad78703e4c8fe703d479c1860d76429e6}.} and removed the crops with an absolute sum of the gradients lower than $1\cdot 10^6$. We used crop sizes of $256\times 256$ pixels, which showed better results for all trained networks. Finally, both augmentation strategies were combined in such a manner that different illumination surfaces were added to each extracted crop.

\subsection{Deep Learning models}

We evaluated and trained six known neural network architectures in order to understand their capabilities for enhancing document images. All networks were trained using two Nvidia RTX 2080 GPUs with 12 GiB of VRAM and implemented using the Keras Framework~\citep{keras} with Tensorflow 1~\citep{tensorflow} as the backend. We will now describe each of these architectures and provide training details and parameters used.

\paragraph{U-Net}
The U-Net architecture was initially proposed to compute semantic segmentation masks for biological images~\citep{unet}. However, because of its success in this task and generalization capabilities, many researchers started to employ it in other domains~\citep{docunet, runet, app_unet, deepotsu}. The design of the U-Net architecture is based essentially on that of an auto-encoder~\citep{autoencoder}.

For the document enhancement task, our initial hypothesis was that this architecture could work by learning the most important visual features of the image (thus eliminating redundant information such as background illumination) in its contracting path (\emph{i.~e.}, the \emph{encoder}), and then reconstructing the image using the compressed features in its expansive path (\emph{i.~e.}, the \emph{decoder}).

We trained the default U-Net architecture with a reduced number of neurons for each layer: $15$, $30$, $60$ and $120$ respectively for both the encoder and decoder segments, due to hardware limitations. The network was trained with the Adam optimizer~\citep{adam}, a learning rate of $5\cdot10^{-2}$, mean squared error (MSE) as the loss function, and batch size of 5 images.

\paragraph{EDSR}
The Enhanced Deep Residual Networks for Single Image Super-Resolution (EDSR) is a model proposed for super resolution tasks~\citep{edsr}. Its architecture is based on (as the authors named it) improved ResNet blocks~\citep{resnet} and skip connections. ResNet blocks are effective known as convolutional blocks, while skip connections usually lead to better and faster convergence of the network. We considered this network for investigation as the feature extraction of ResNet blocks and the propagation of these features via skip connections were shown to be promising for preserving document content during initial tests.

We trained the default architecture\footnote{We used the following EDSR base model implementation: \url{https://github.com/krasserm/super-resolution/blob/master/model/edsr.py}.} with $8$ improved ResNet blocks containing $64$ neurons in each convolutional layer. However, we removed the up-sample blocks, since we aimed to maintain the input shape and leverage only their feature extraction module. The network was trained with the Adam optimizer~\citep{adam}, a learning rate of $1\cdot10^{-6}$, MSE as the loss function, and a batch size of $22$ images.

\paragraph{WDSR}
The Wide Activation for Efficient and Accurate Image Super-Resolution (WDSR) is another model initially proposed for super resolution tasks~\citep{wdsr}. Like EDSR, this architecture is also based on ResNet blocks~\citep{resnet} and skip connections. However, the authors proposed a pre-expansion of the features before the activation blocks and  replacement of the batch normalization layers by an optimizer that uses weight normalization instead~\citep{weights_normalization}.

In the original paper, the authors introduced two architectures, namely WDSR-A and WDSR-B. The major difference between these is the size of the expansion block. While  WDSR-A augments from $2$ to $4$ times, WDSR-B augments from $6$ to $9$ times the convolutional neurons before the activation layers. This is used to achieve different image augmentations on the original super resolution task, but it is not relevant for this work.

We trained the default architectures\footnote{We used the following WDSR base model implementation: \url{https://github.com/krasserm/super-resolution/blob/master/model/wdsr.py}.} with $8$ ResNet blocks containing $32$ neurons in the convolutional layers, $4$-time expansion in the WDSR-A architecture and a $6$-time expansion in the WDSR-B. However, we removed the up-sample blocks for the same reason as in the EDSR architecture. The networks were trained using a Weight Normalization Adam optimizer\footnote{We used the following Weight Normalization implementation: \url{https://github.com/openai/weightnorm/blob/master/keras/weightnorm.py}.} implementation~\citep{weightnorm}, a learning rate of $1\cdot10^{-6}$, MSE as the loss function, and a batch size of $15$ images.

\paragraph{SRGAN}
The SRGAN architecture relies on a Generative Adversarial Network (GAN) trained
with feature loss~\citep{srgan}. The feature loss uses the last convolutional
layer (the backbone output) from a VGG~\citep{vgg} network trained on the ImageNet dataset~\citep{imagenet}. The goal of this architecture is to generate reliable super-resolution images from its low resolution form, without using a loss function based on pixel-wise comparisons between the network output and a ground truth. This is achieved by using two concurrent neural networks during training: the \emph{generator}, which transforms an input image into another image of the same size; and the \emph{discriminator}, which is essentially a classifier to determine whether the output image from the generator is a real (ground truth) image or a generated (\emph{i.~e.}, synthetic) image. This is the basic principle of training Generative Adversarial Neural Networks~\citep{gans}.

In order to train this network for document enhancement, we replaced the default VGG16~\citep{vgg} feature extractor model by an internal pre-trained model that classifies different types of documents using the MobileNet V1~\citep{mobilenetv1} as the backbone. The expected result of this modification was to improve the feature loss propagation, since this internal model already was trained to extract features from document images (instead of natural images as pre-trained VGG on Imagenet). The up-sample blocks were also removed, for the same reasons as discussed for EDSR and WDSR.

For our use case, the generator network receives a raw document image as input and the discriminator classifies whether the generator's output was an enhanced document or not, based on examples of actual enhanced documents from our training set. Initially, the generator was pre-trained using a pixel-wise loss function with ground truth images (similarly to the other models) for $3$ epochs with $1000$ iterations per epoch. In this pre-training step, the Adam optimizer~\citep{adam} was used with a fixed learning rate of $1\cdot10^{-6}$ and MSE as the loss function. Then, we continue by training the entire GAN network\footnote{The SRGAN training used the same hyper-parameters and procedures as presented in the following repository: \url{https://github.com/eriklindernoren/Keras-GAN\#srgan}.}. To evaluate the evolution of GAN training, the generator was evaluated after each $1000$ iterations with the WaDIQaM metric~\citep{wadiqam}, saving the results every time the network improved.

\paragraph{IRCNN}
The IRCNN~\citep{ircnn} is a denoiser model~\citep{img_denoiser}, which receives a noisy image as input and estimates a residual image in its output. This residual image can then be subtracted from the noisy input to return the clean (\emph{i.~e.}, enhanced) image. For document enhancement, the goal is to produce a residual image that is a close estimate of the illumination surface of the document image, along with other undesired artifacts.

We used the default hyper-parameters for the architecture\footnote{We used the following IRCNN base model implementation: \url{https://github.com/lipengFu/IRCNN}.} and added a subtraction layer at the end of the model, which receives the input image and the estimated residual image, resulting in the enhanced document image. For training, we used the Adam optimizer~\citep{adam}, a learning rate of $1\cdot10^{-4}$, MSE as the loss function, and a batch size of $20$ images.

\paragraph{CycleGAN}
The CycleGAN network~\citep{cyclegan} is an unpaired image-to-image translator~\citep{image_translation}. This is achieved by training two generators and discriminators using the same basic process used for train generative adversarial networks~\citep{gans}.

For document enhancement, the idea is to train this network for converting a raw image into an enhanced one, and vice-versa. A useful side-effect of this process is that the reverse conversion (from enhanced document back to captured) may potentially be re-used for data augmentation later.

We trained the model using default architecture and training hyper-parameters from an existing implementation\footnote{We used the same CycleGAN hyper-parameters as presented in the following repository: \url{https://github.com/eriklindernoren/Keras-GAN\#cyclegan}.}. The raw-to-enhanced generator was evaluated after each $1000$ iterations with the WaDIQaM~\citep{wadiqam} metric, saving the results every time the network improved. However, we observed
that this model is overly sensitive to the weights update, and often gets stuck in local minima. To address this, after the evaluation, if the model did not improve, the weights from the last best evaluation were re-applied on the model.

\subsection{Evaluation} \label{sec:iqa_methodology}
To evaluate the models, we used a two-step approach: 
visual inspection and a numeric comparison using IQA metrics. First, we enhanced all the document images from our test set using the trained models. Then, we manually inspected the enhanced images in order to validate whether the enhanced images were computed correctly when compared to the ground truth. During inspection, we focused on the following aspects of the returned enhanced image: correctness of illumination surface removal, content preservation, amount of contrast between content and background, and color accuracy. This step was necessary, as none of the models or IQA metrics tested were specifically designed to provide robust values for assessing document quality in particular. Therefore, we avoided further comparisons for cases where the metrics could fail to accurately reflect the perceived document quality.

Next, to compare results of models that provided a visual good enhancement, we used the peak signal-to-noise ratio (PSNR), the multi-scale structural similarity (MS-SSIM)~\citep{ssim}, the Perceptual Image Error (PIE)~\citep{pie} and the Weighted Average Deep Image Quality Measure (WaDIQaM)~\citep{wadiqam}. For more information regarding the methodology used to choose these metrics, please refer to the Appendix~\ref{sec:iqa_investigation}.

The PSNR is defined as:
\begin{equation}
	\mathrm{PSNR} = 20 \cdot \log_{10}{\left(\max_{p\in I}{p}\right)} - 10 \cdot \log_{10}{\left(\mathrm{MSE}\right)} ,
\end{equation}
where $\max_{p\in I}{p}$ is the maximum possible pixel value of the image (\emph{e.~g.}, $255$ for 8-bit images) and $\mathrm{MSE}$ is the mean squared error. The PSNR measures the ratio between the maximum possible power of a signal and the power of the corrupting noise. 

MS-SSIM~\citep{ssim} is an IQA metric that evaluates the perceived change in the structural information between two images. It uses the SSIM index, which is calculated using the mean, variance and covariance values of various windows of the compared images\footnote{We used the following MS-SSIM implementation, available on the scikit-image Python library: \url{https://scikit-image.org/docs/dev/api/skimage.metrics.html?highlight=ssim\#skimage.metrics.structural_similarity}.}.

The PIE~\citep{pie}\footnote{We used the following official PIE implementation: \url{https://github.com/prashnani/PerceptualImageError}.} and WaDIQaM~\citep{wadiqam}\footnote{We used the following non-official WaDIQaM implementation: \url{https://github.com/lidq92/WaDIQaM}.} are deep learning based metrics that were trained to return scores that attempts to match those as perceived by humans. Both methods use a similar strategy for scoring and weighting patches of the pairwise input images to return the final score; however, they differ in the methodology used to define their labels during training (which, in practice, result in different evaluation scores, measuring different aspects of compared images).

\section{Results and discussion}\label{sec:results}

In this section, we present and discuss the results from the evaluation methodology proposed in Section~\ref{sec:experimental-setup}. We will first show the results from the networks that generally performed poorly for most images (\emph{i.~e.}, \emph{failed} networks) and briefly discuss these results. Then, results from the remaining networks (\emph{i.~e.}, networks considered to \emph{pass} the qualitative evaluation) will be presented next.

\begin{figure}[htb]\centering
	\subfloat[Ground truth]{\includegraphics[width=0.2\textwidth]{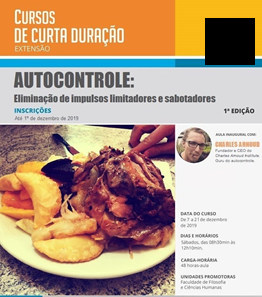}}
	~
	\subfloat[WDSR-B]{\includegraphics[width=0.2\textwidth]{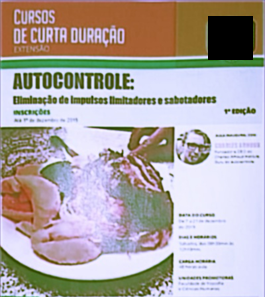}}
	~
	\subfloat[IRCNN]{\includegraphics[width=0.2\textwidth]{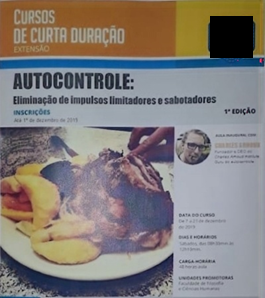}}
	~
	\subfloat[CycleGAN]{\includegraphics[width=0.2\textwidth]{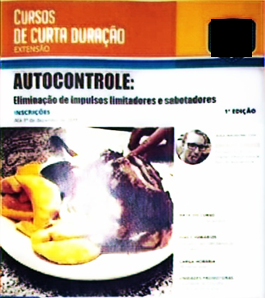}}
	\caption{Example of network results that were not considered to be accurately enhanced from manual inspection.}
	\label{fig:failed}
\end{figure}

The following networks generally failed to achieve visually satisfactory results:

\begin{description}
    \item[WDSR-B] generally failed to perform lighting correction, sharpening the textual content with relation to the background and preserving color accuracy;
    \item[IRCNN] failed to perform lighting correction in general and sharpening the textual content; and
    \item[CycleGAN] erased blurry or low contrast content in general. Figure~\ref{fig:failed} compares the output of an image from each network with its ground truth.
\end{description}

\begin{figure}[htb]\centering
	\subfloat[Raw image]{\includegraphics[width=0.2\textwidth]{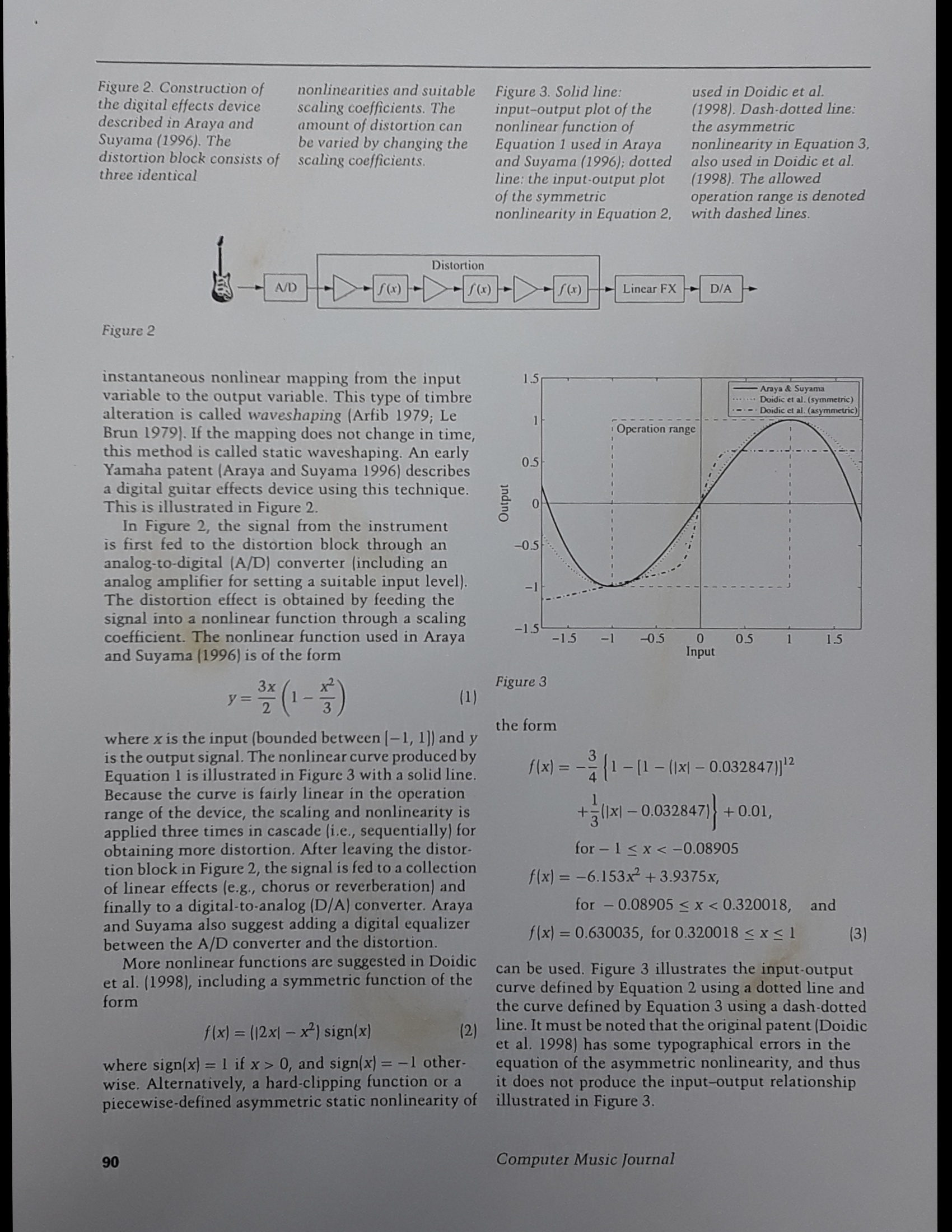}}~
	\subfloat[Ground truth]{\includegraphics[width=0.2\textwidth]{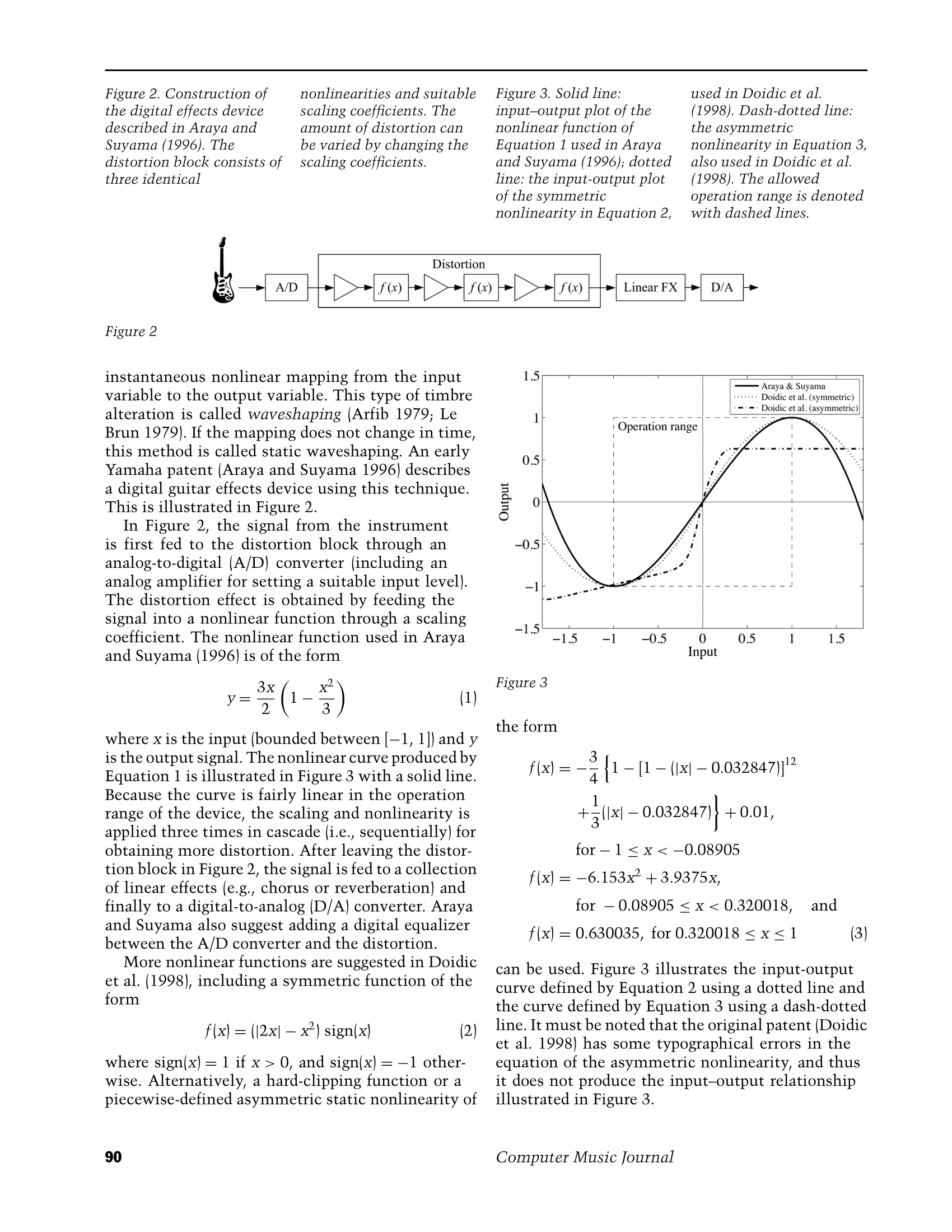}}~
	\subfloat[\cite{pagelift}]{\includegraphics[width=0.2\textwidth]{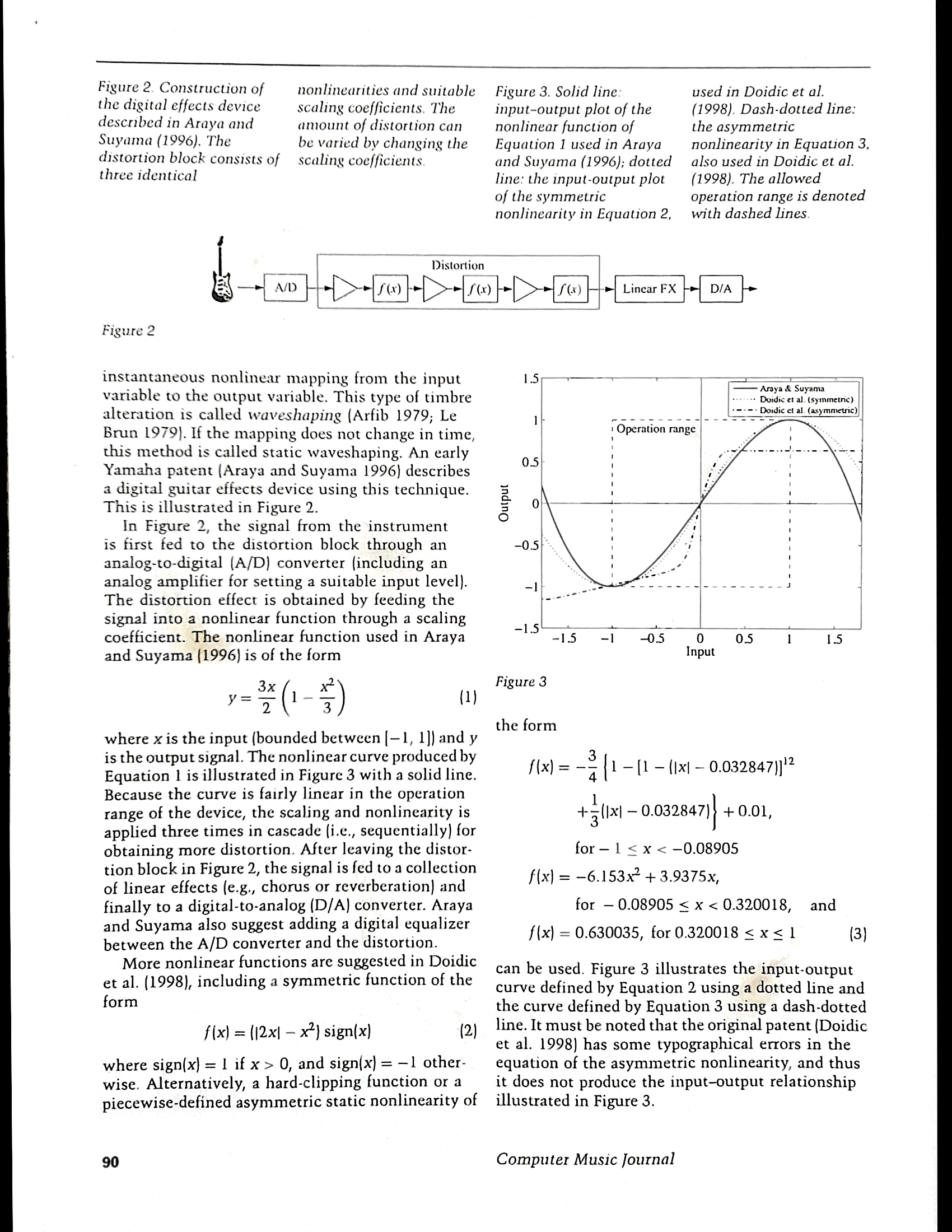}}
	\\
	\subfloat[EDSR]{\includegraphics[width=0.2\textwidth]{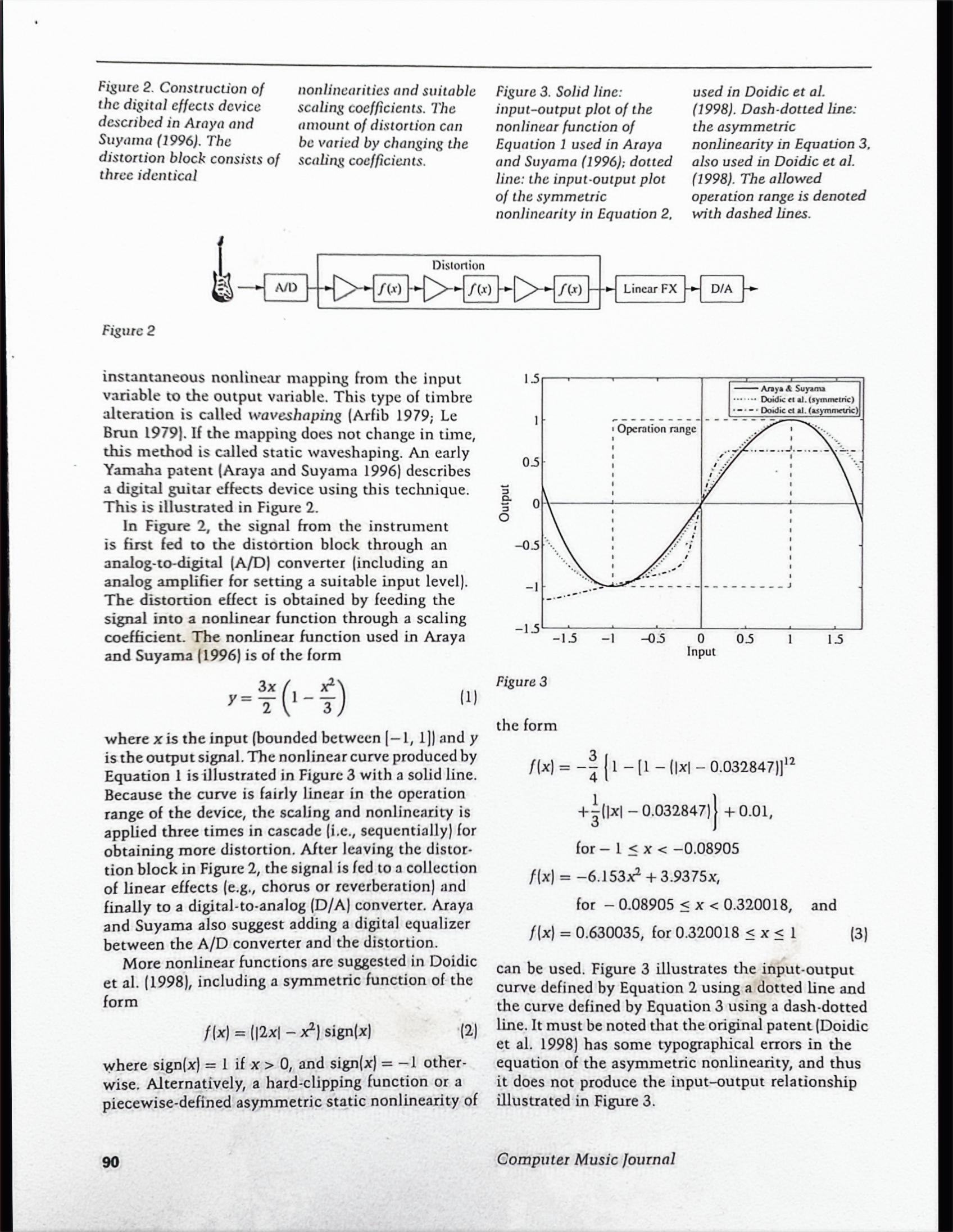}}~
	\subfloat[SRGAN]{\includegraphics[width=0.2\textwidth]{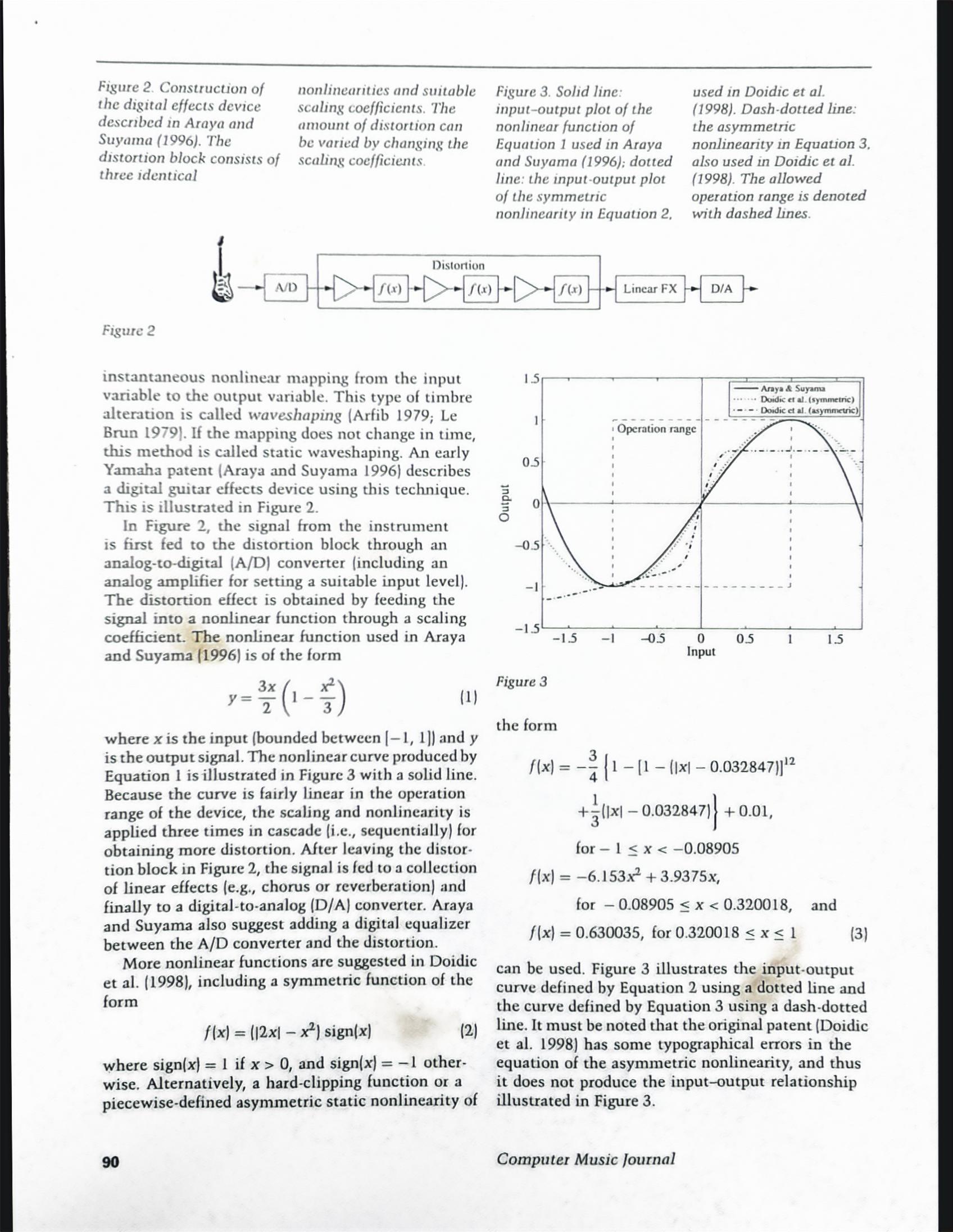}}~
	\subfloat[WDSR-A]{\includegraphics[width=0.2\textwidth]{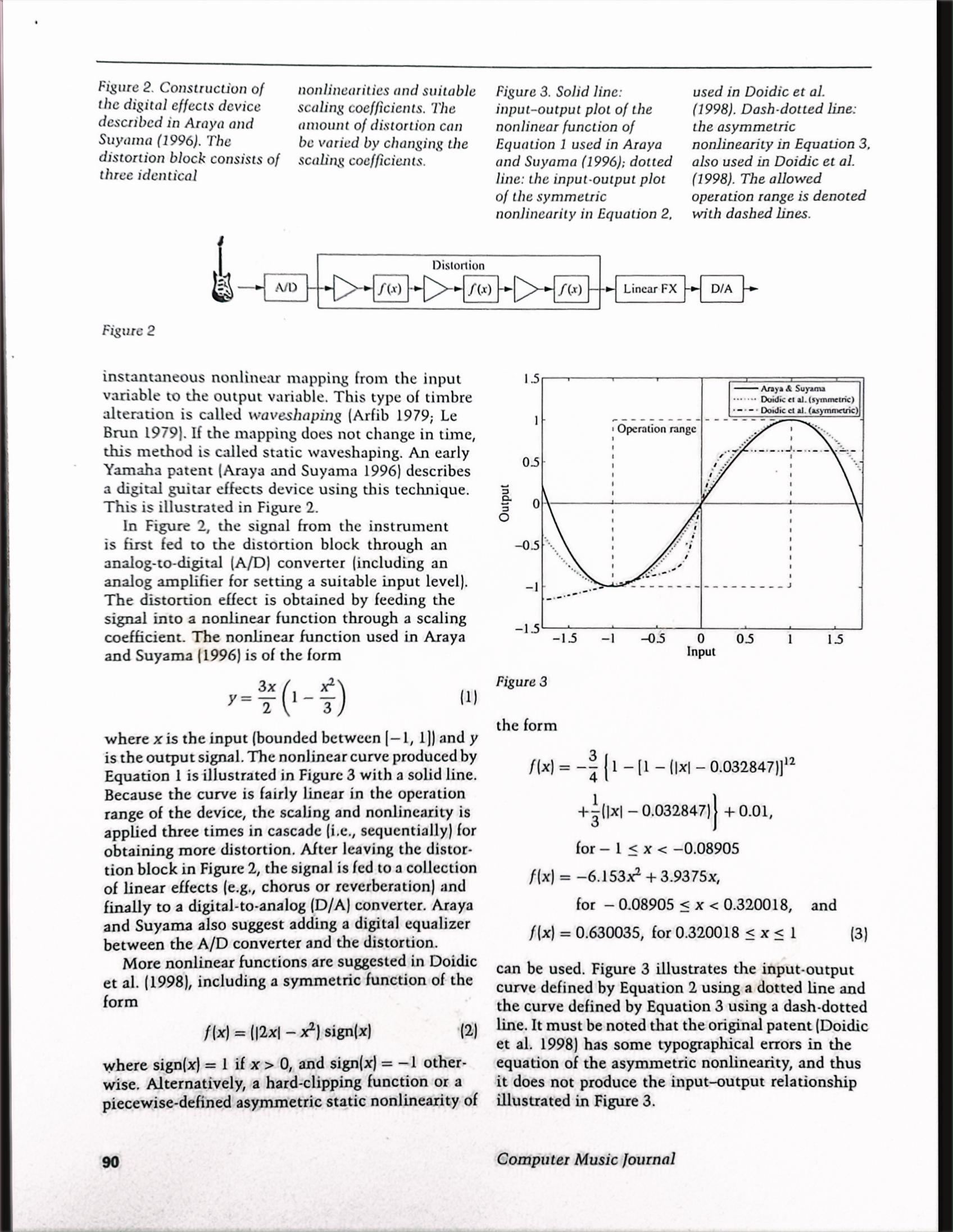}}~
	\subfloat[U-Net]{\includegraphics[width=0.2\textwidth]{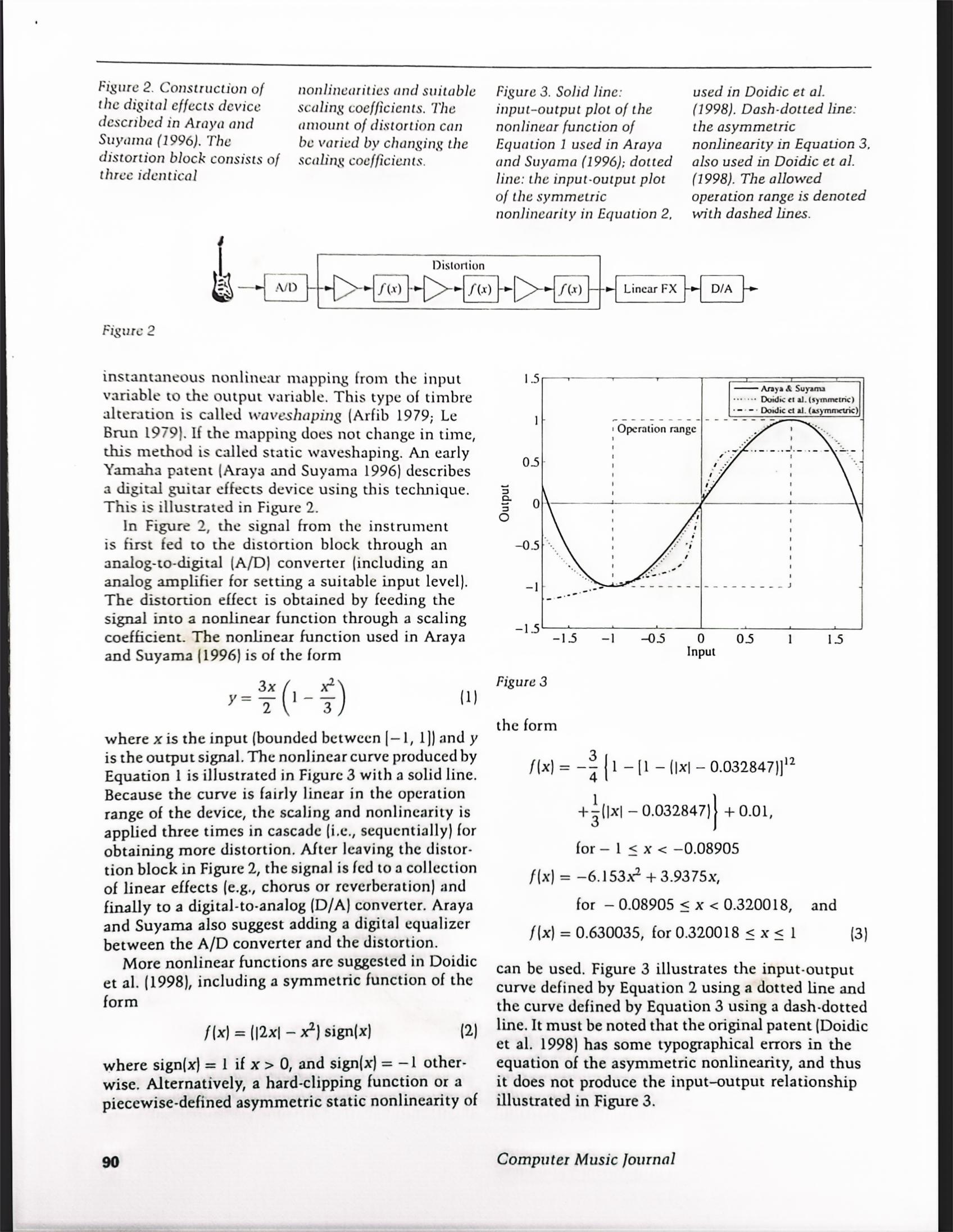}}
	\caption{Results for a black-and-white document image}
	\label{fig:good_1}
\end{figure}

\begin{figure}[htb]\centering
	\subfloat[Raw image]{\includegraphics[width=0.2\textwidth]{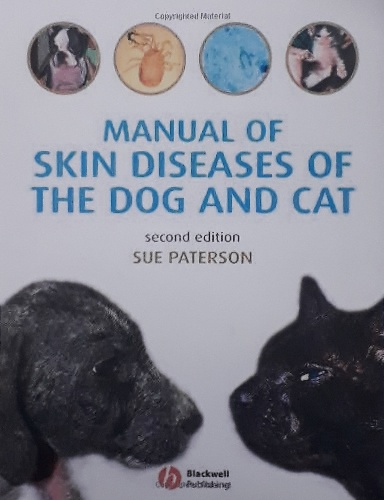}}~
	\subfloat[Ground truth]{\includegraphics[width=0.2\textwidth]{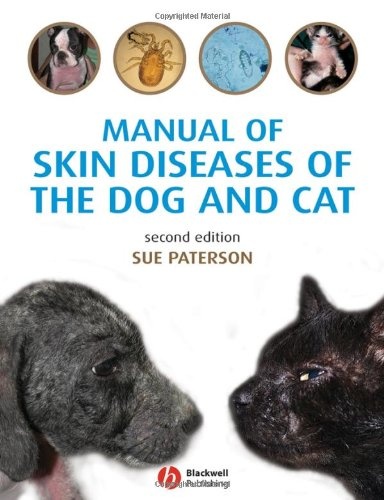}}~
	\subfloat[\cite{pagelift}]{\includegraphics[width=0.2\textwidth]{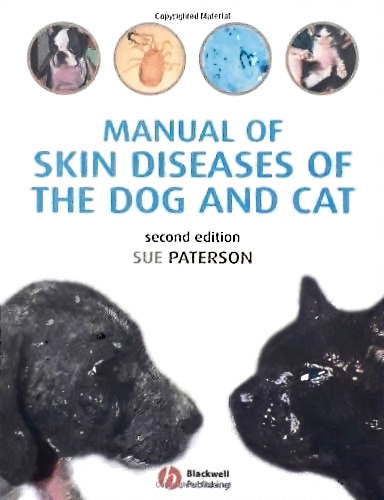}}
	\\
	\subfloat[EDSR]{\includegraphics[width=0.2\textwidth]{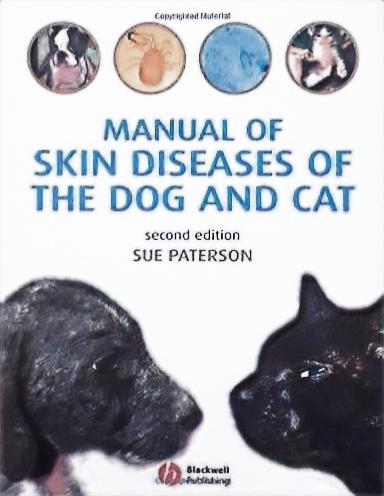}}~
	\subfloat[SRGAN]{\includegraphics[width=0.2\textwidth]{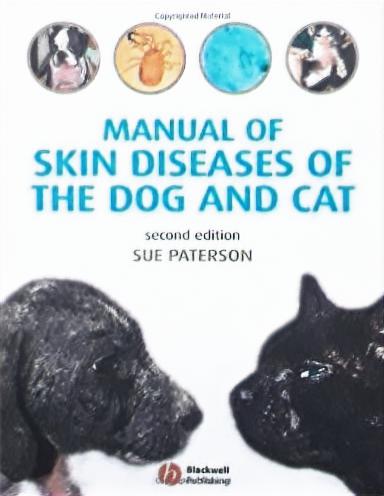}}~
	\subfloat[WDSR-A]{\includegraphics[width=0.2\textwidth]{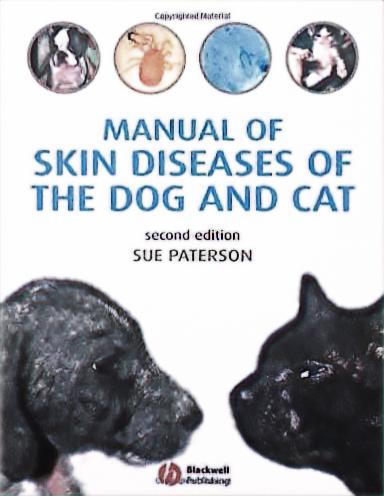}}~
	\subfloat[U-Net]{\includegraphics[width=0.2\textwidth]{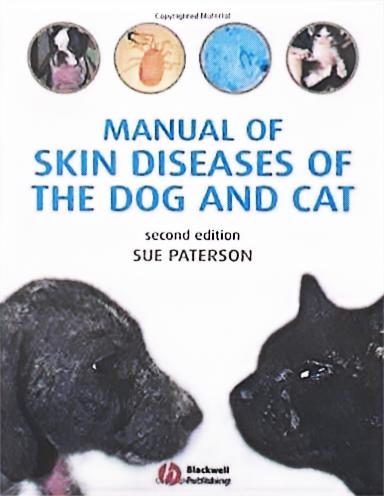}}
	\caption{Results for a colored mixed-content document image}
	\label{fig:good_2}
\end{figure}

\begin{figure}[htb]\centering
	\subfloat[Raw image]{\includegraphics[width=0.2\textwidth]{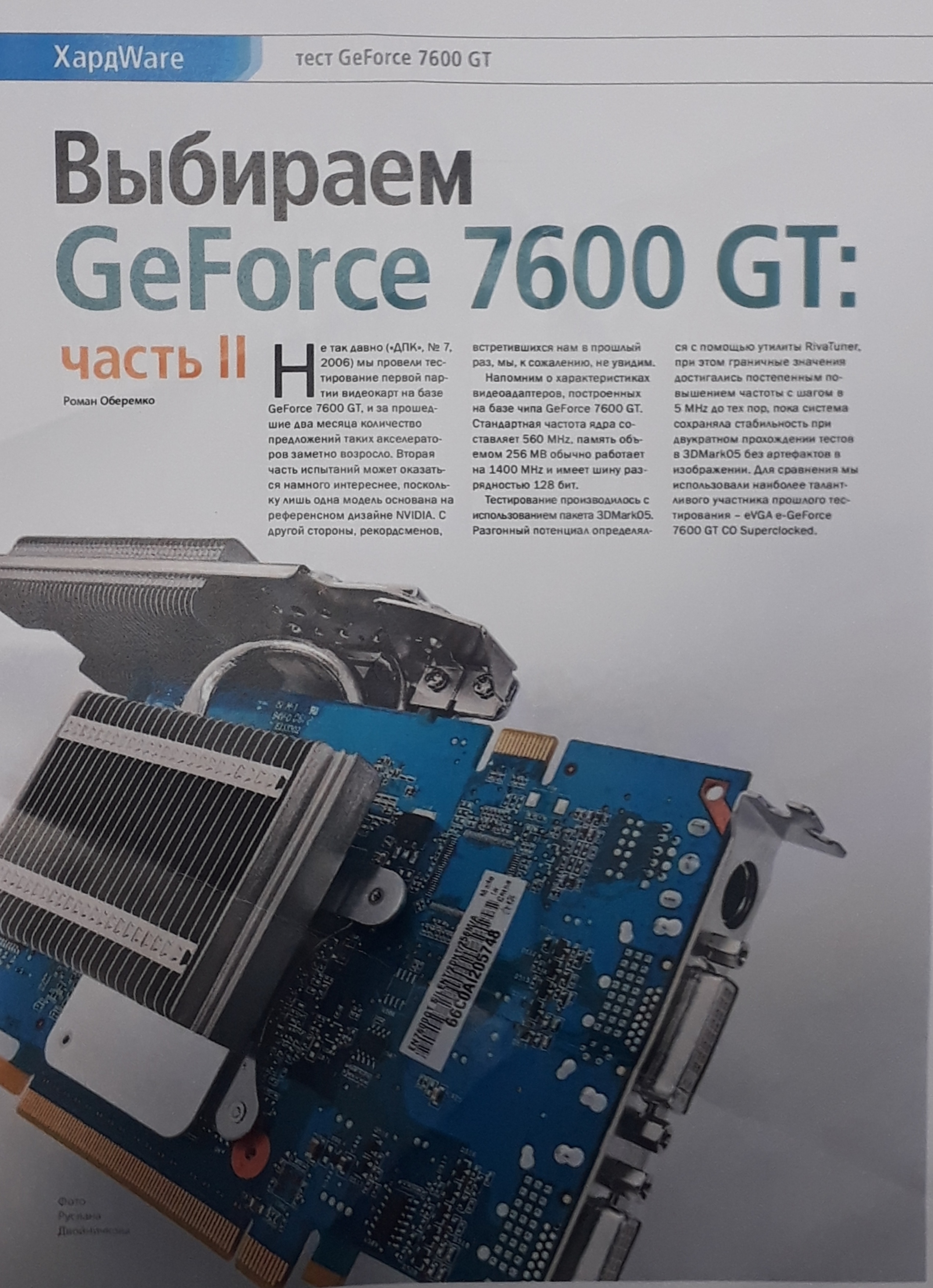}}~
	\subfloat[Ground truth]{\includegraphics[width=0.2\textwidth]{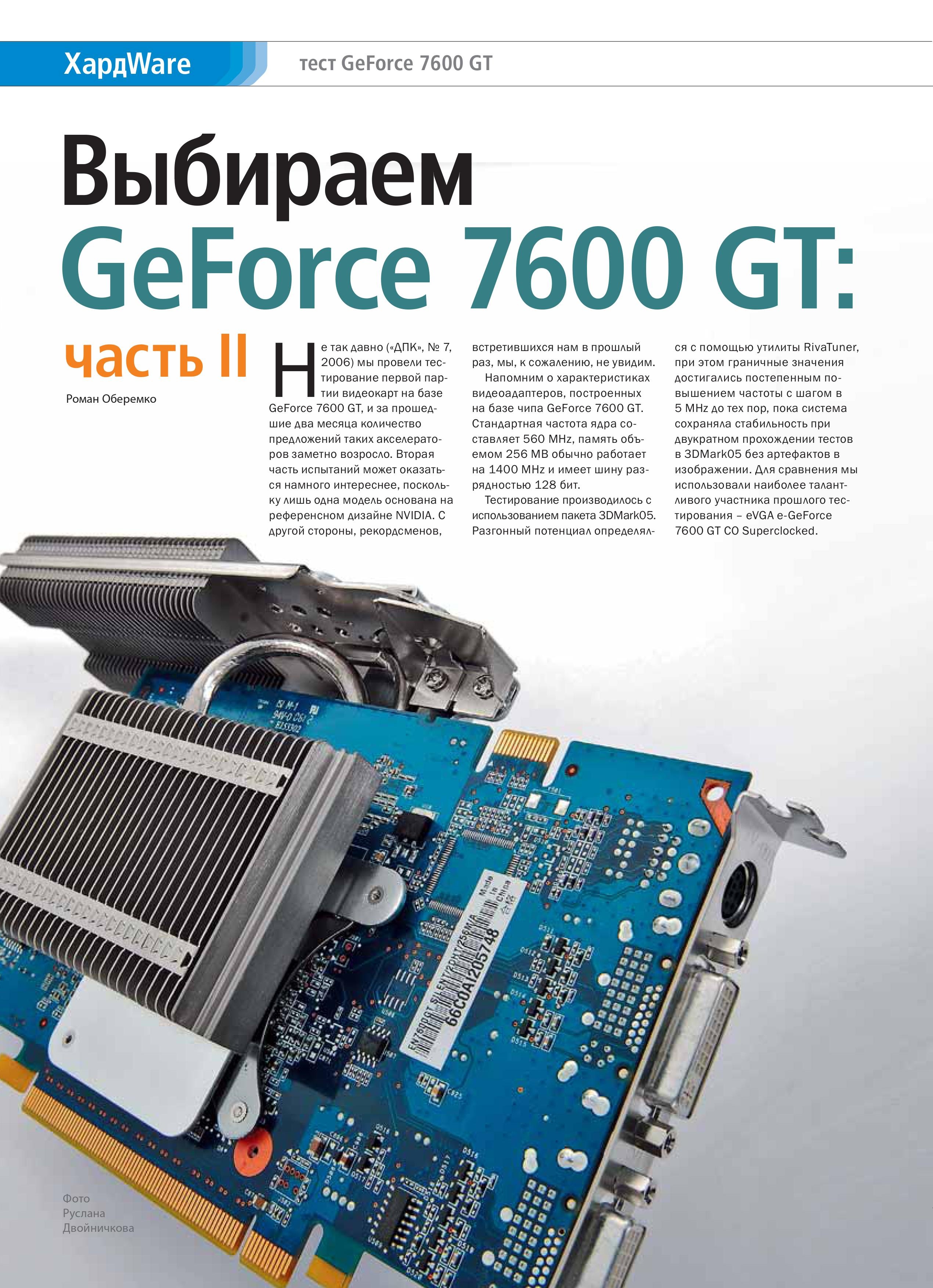}}~
	\subfloat[\cite{pagelift}]{\includegraphics[width=0.2\textwidth]{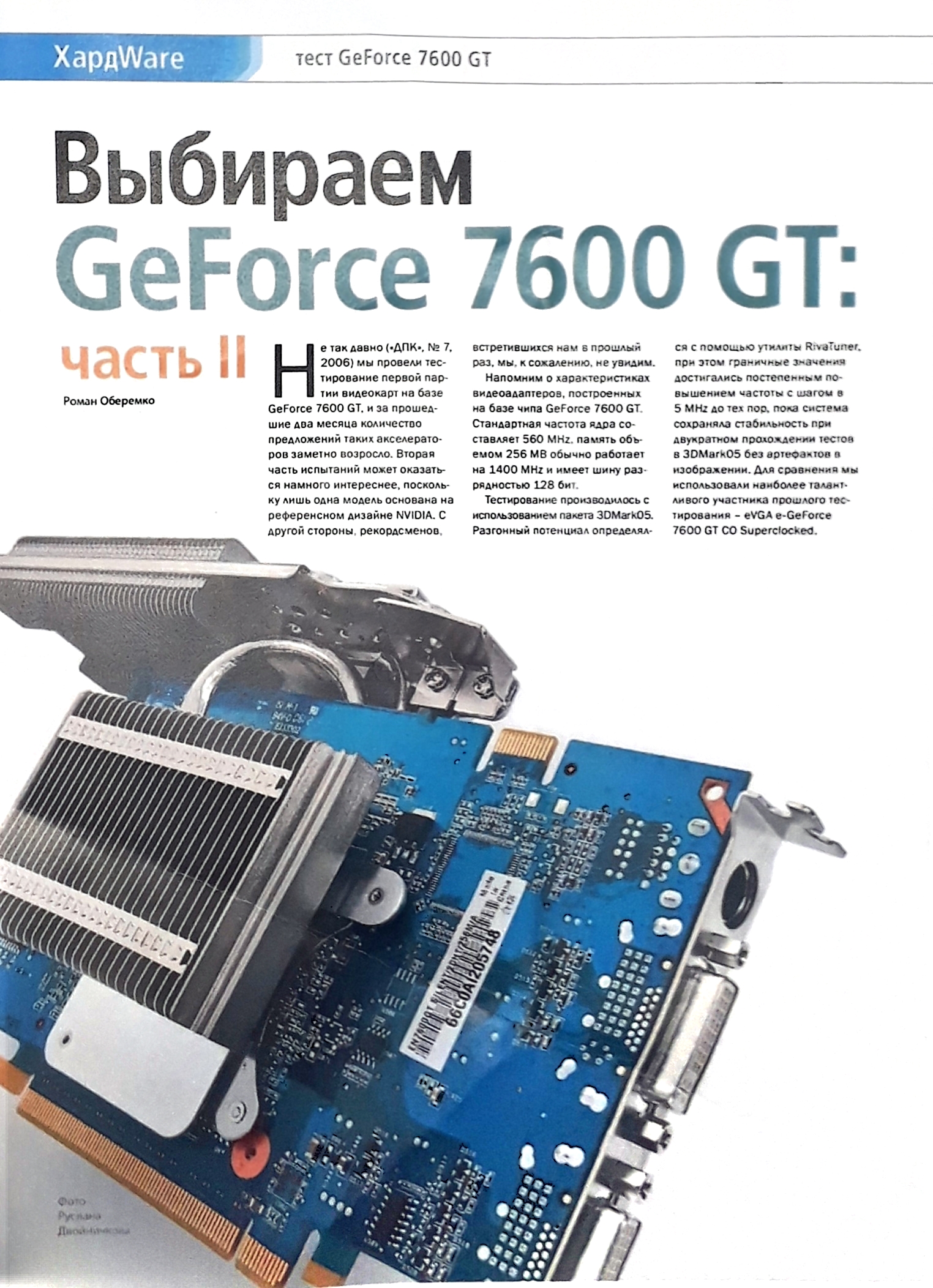}}
	\\
	\subfloat[EDSR]{\includegraphics[width=0.2\textwidth]{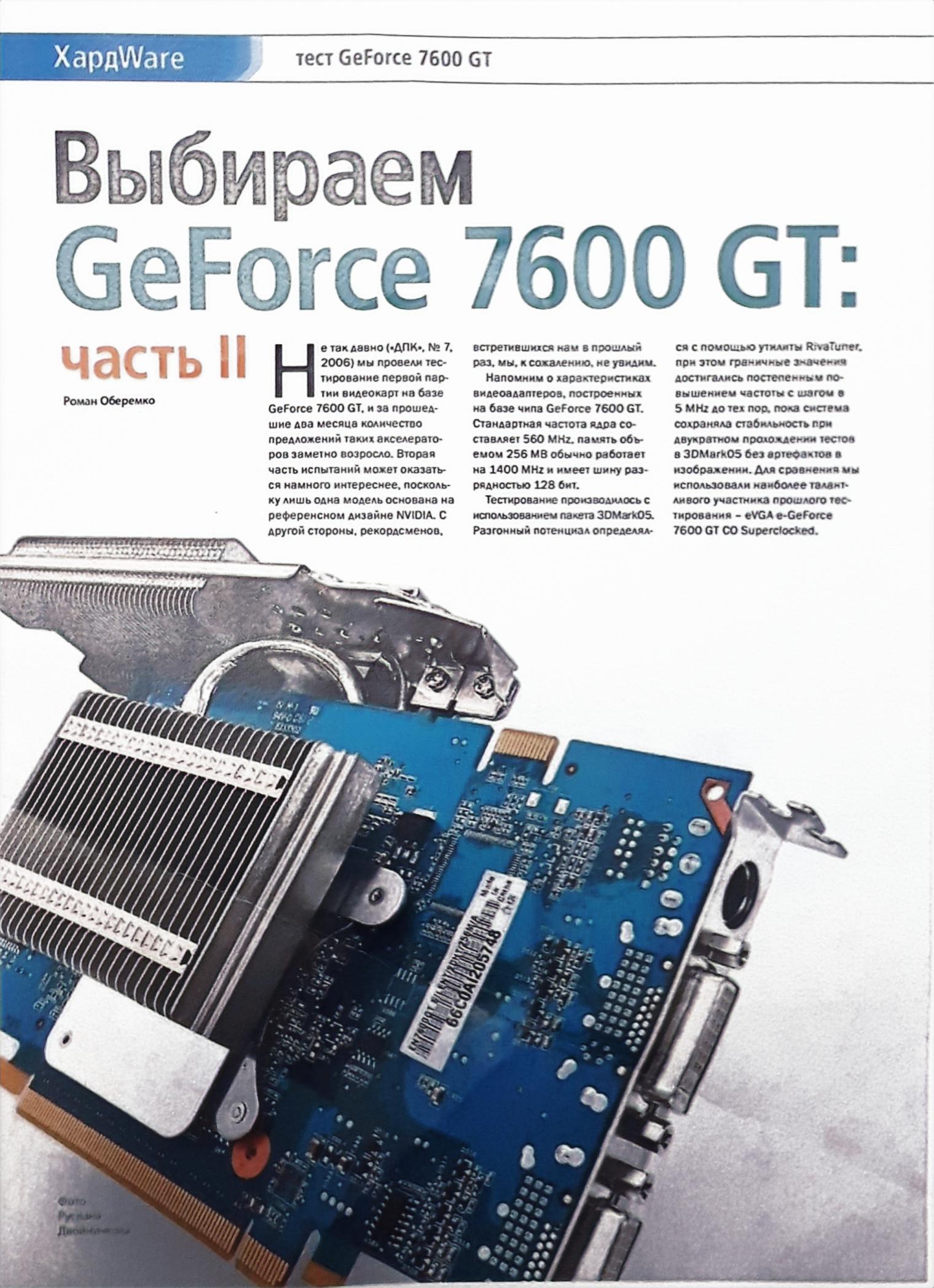}}~
	\subfloat[SRGAN]{\includegraphics[width=0.2\textwidth]{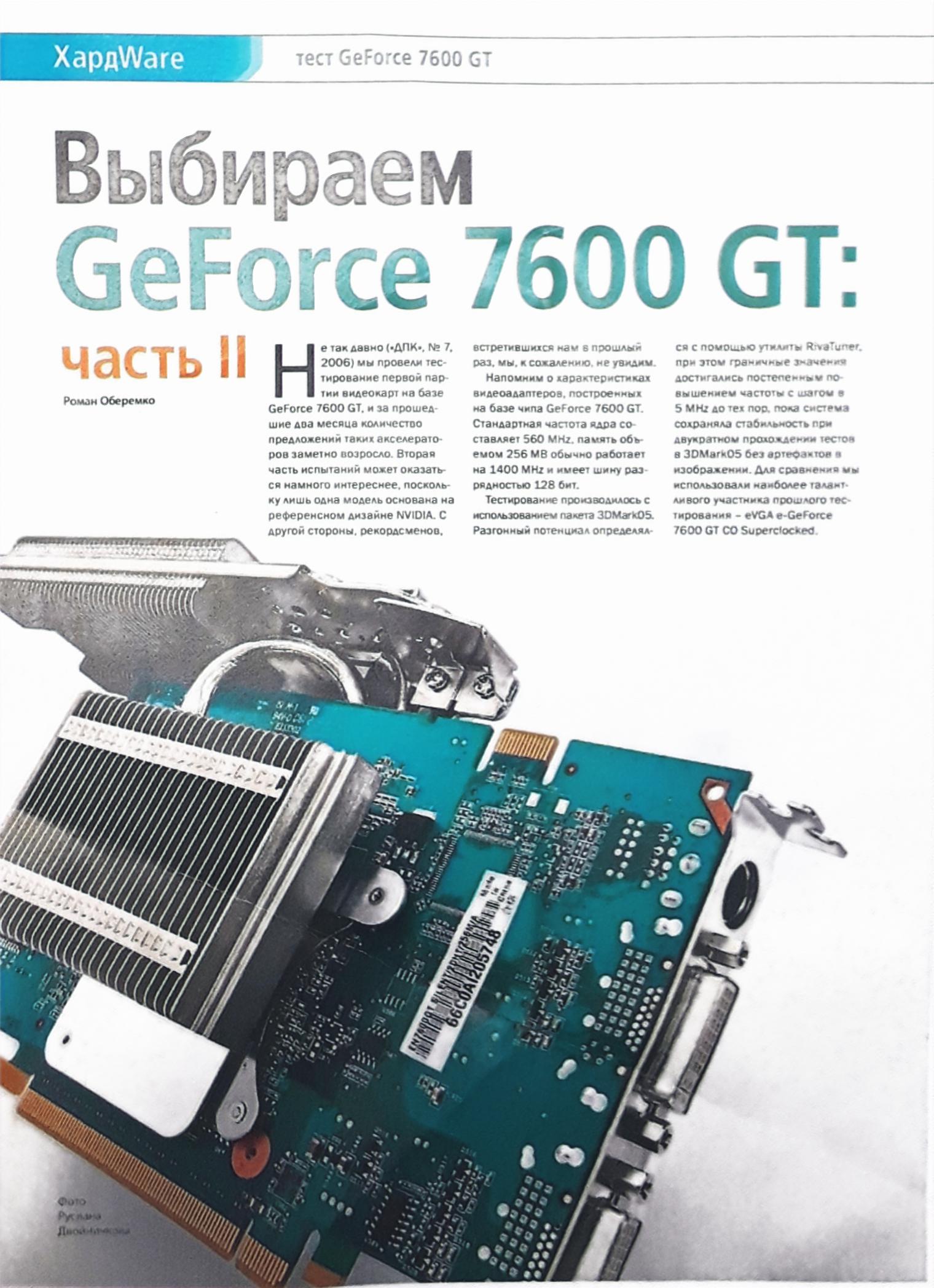}}~
	\subfloat[WDSR-A]{\includegraphics[width=0.2\textwidth]{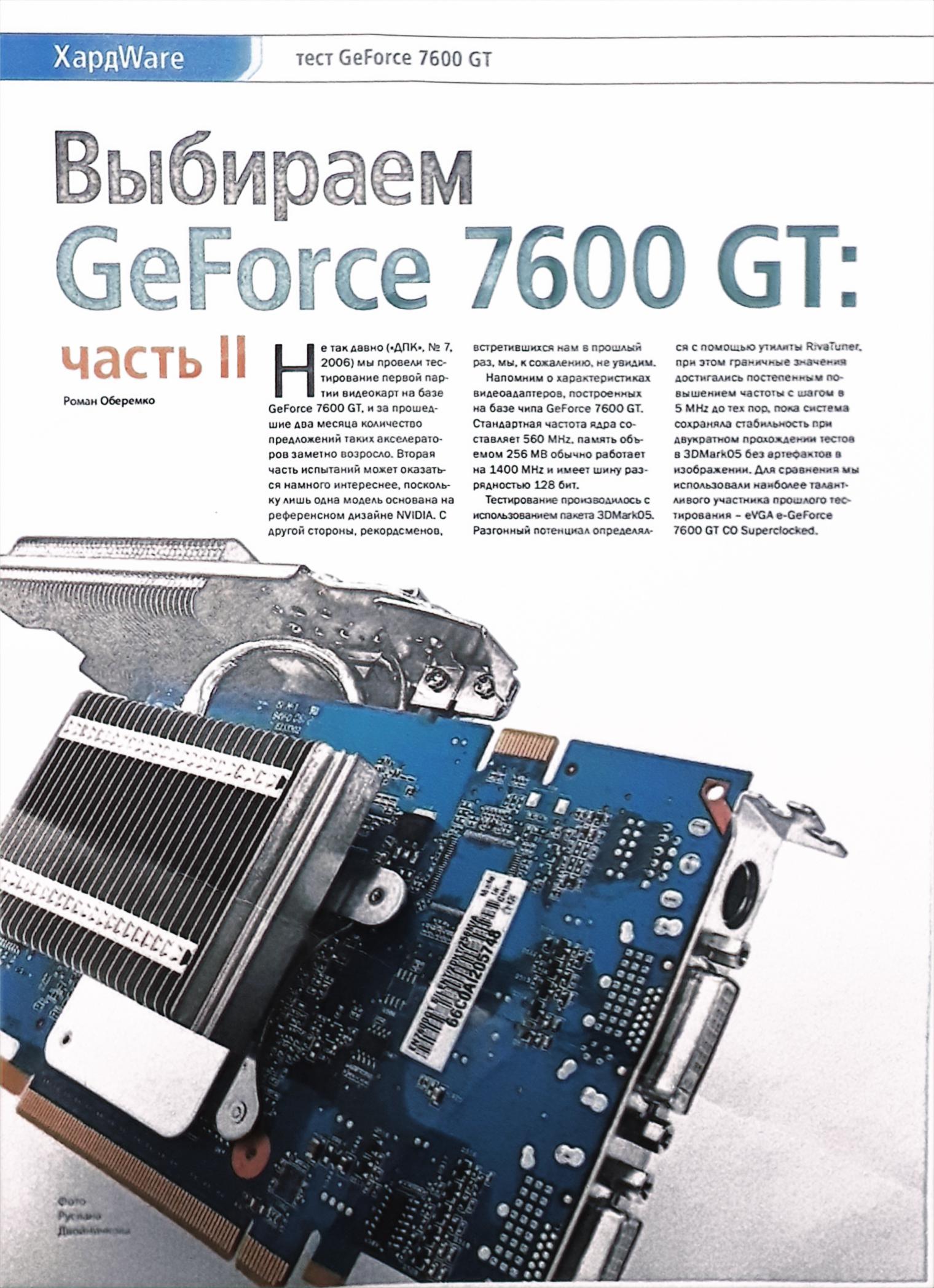}}~
	\subfloat[U-Net]{\includegraphics[width=0.2\textwidth]{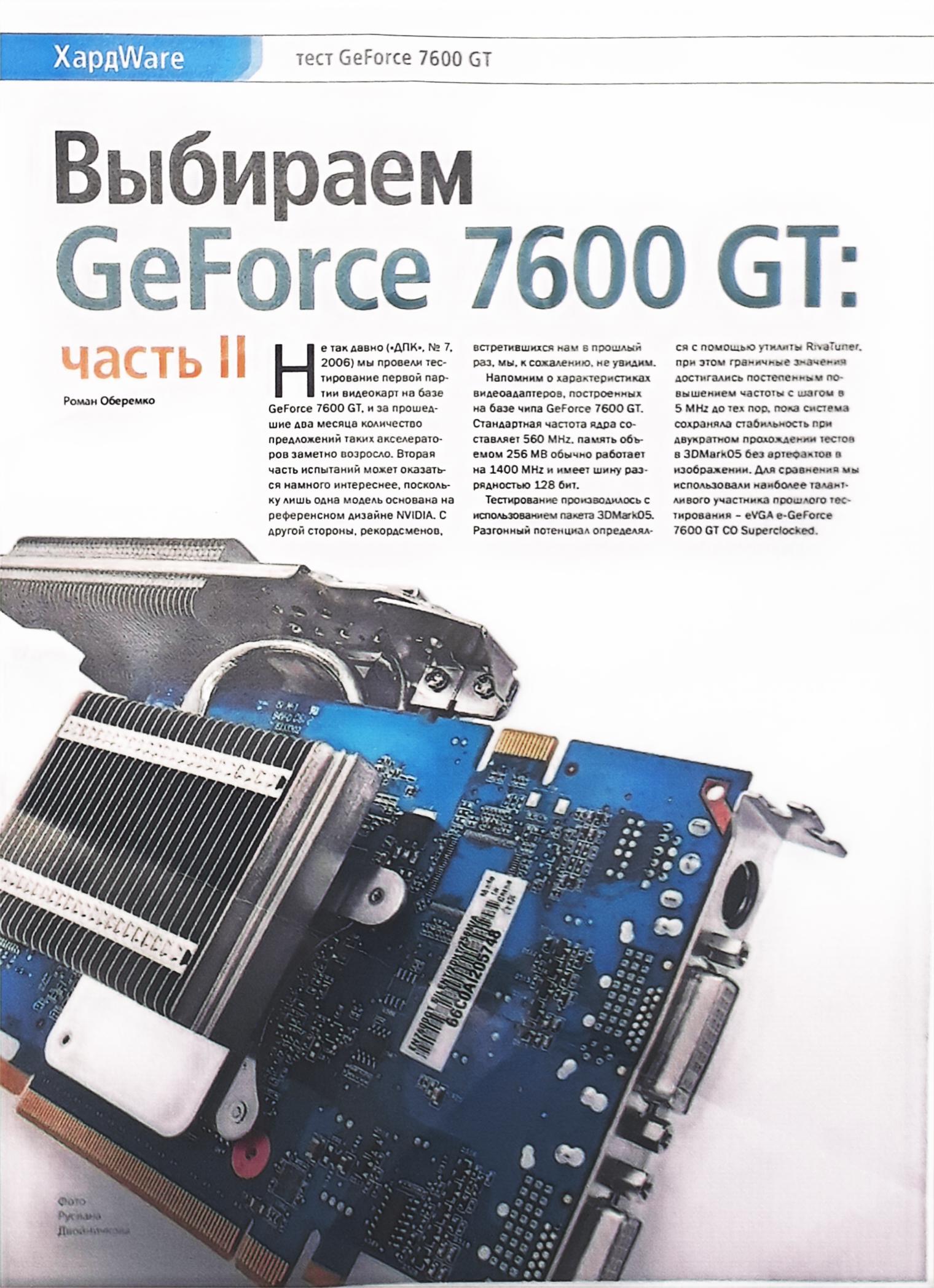}}
	\caption{Results for a mixed-content document image containing ambiguous content (\emph{i.~e.}, shadows)}
	\label{fig:good_3}
\end{figure}

\begin{figure}[htb]\centering
	\subfloat[Raw image]{\includegraphics[width=0.2\textwidth]{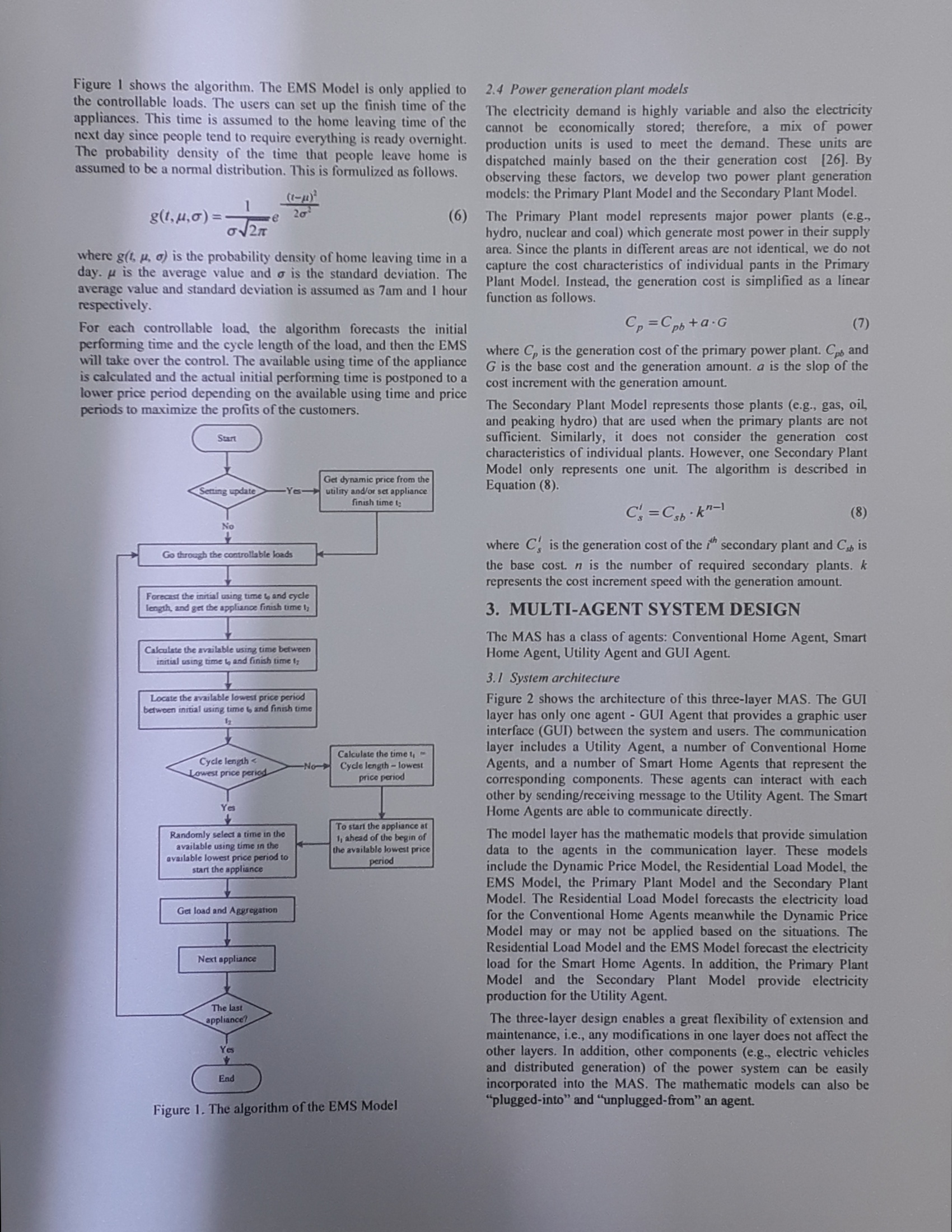}}~
	\subfloat[Ground truth]{\includegraphics[width=0.2\textwidth]{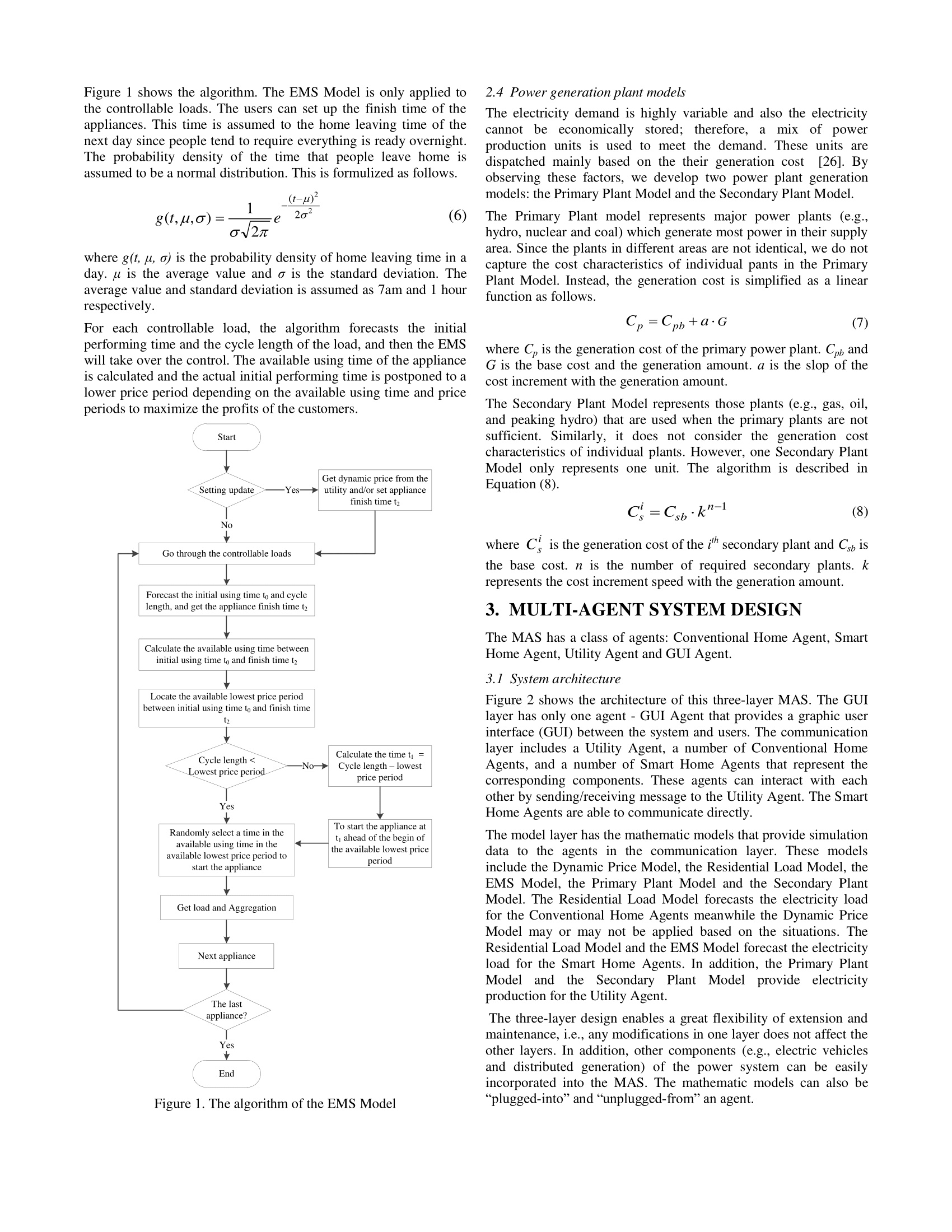}}~
	\subfloat[\cite{pagelift}]{\includegraphics[width=0.2\textwidth]{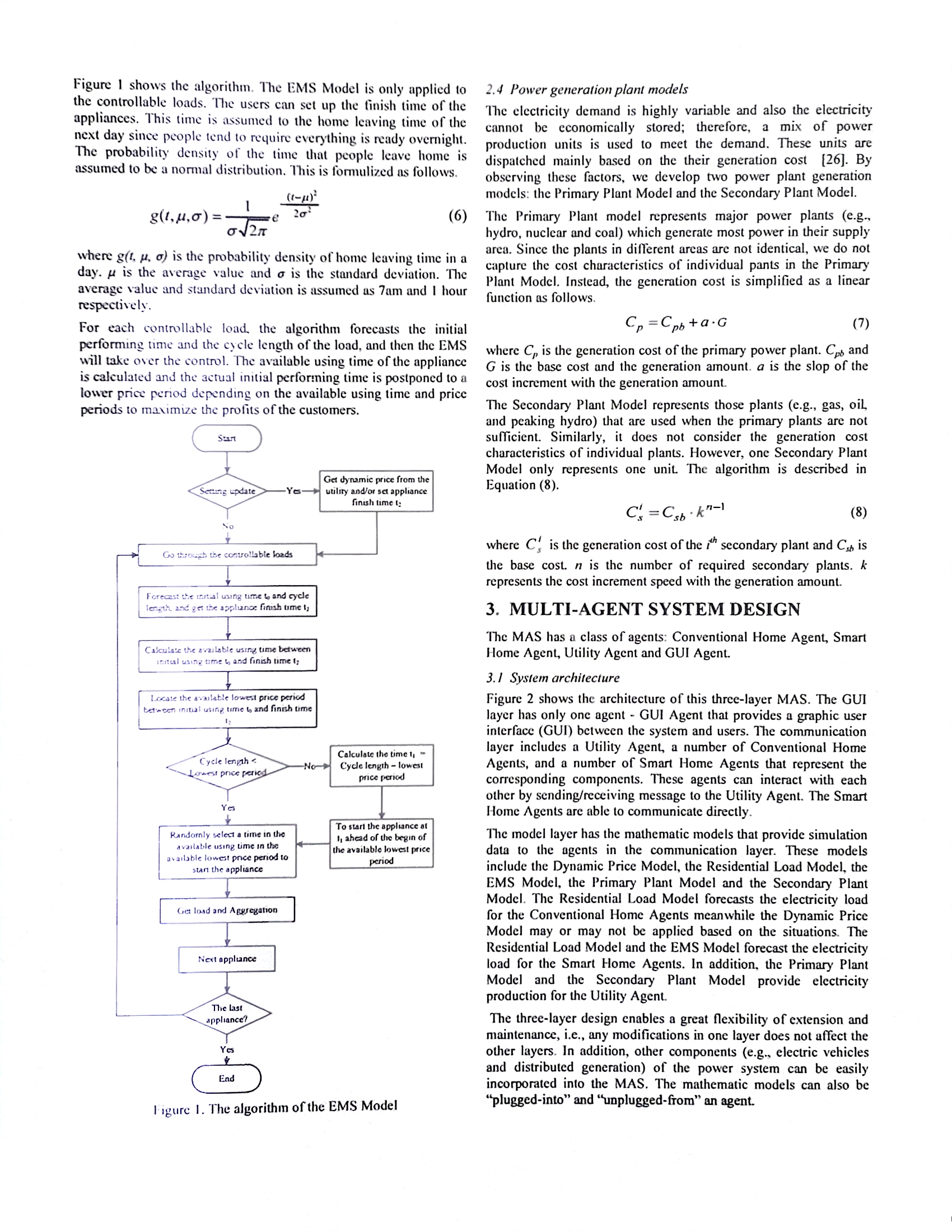}}
	\\
	\subfloat[EDSR]{\includegraphics[width=0.2\textwidth]{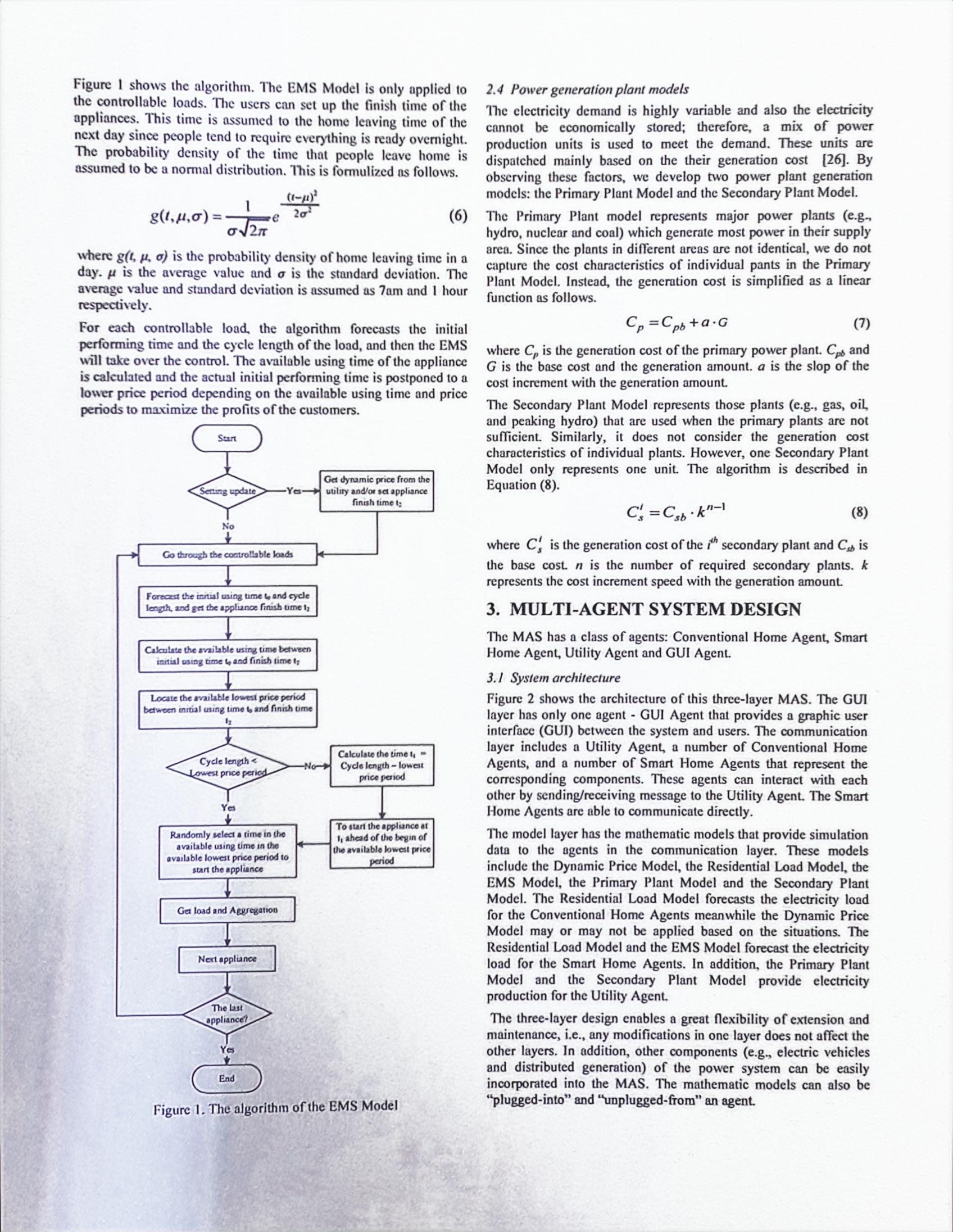}}~
	\subfloat[SRGAN]{\includegraphics[width=0.2\textwidth]{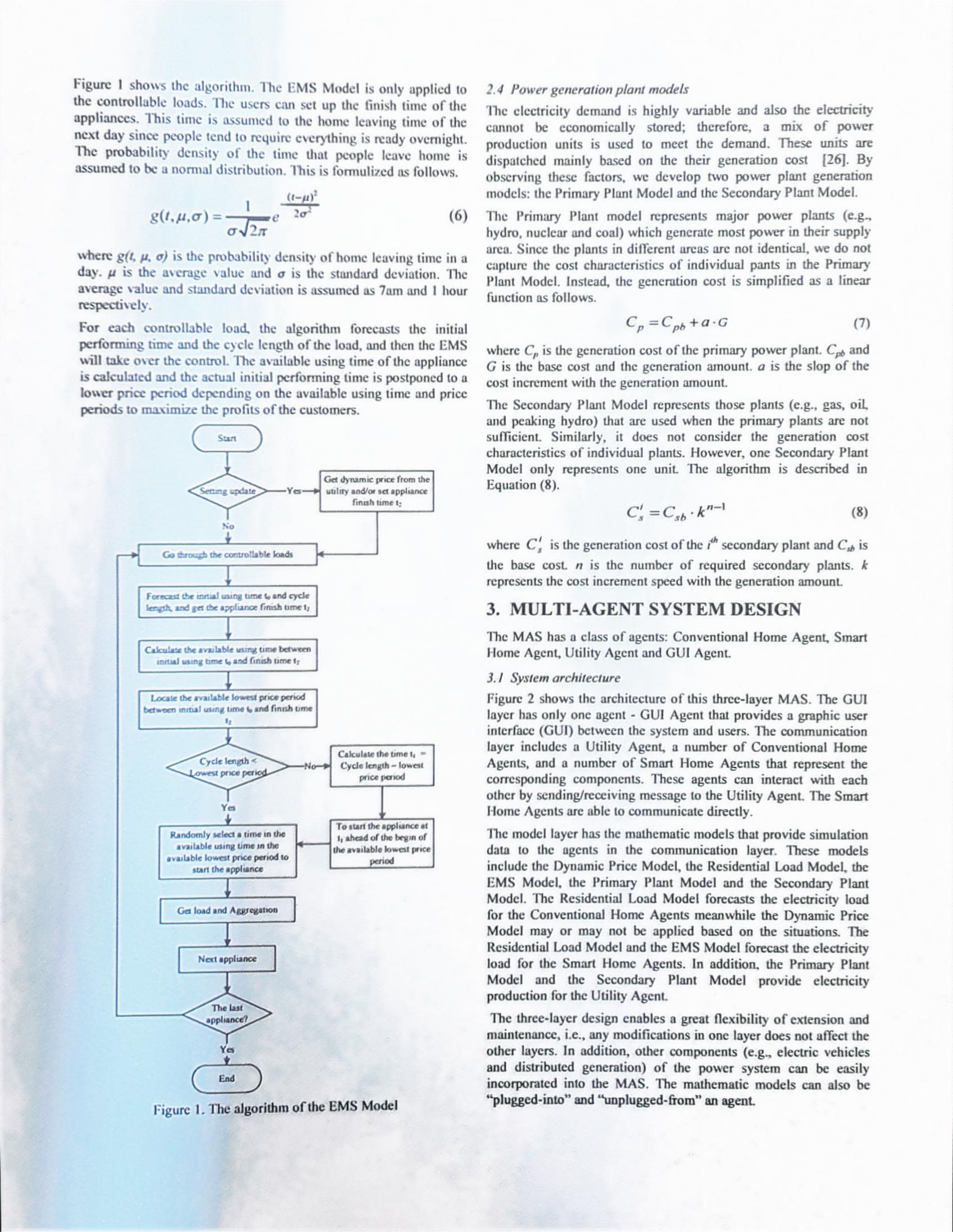}}~
	\subfloat[WDSR-A]{\includegraphics[width=0.2\textwidth]{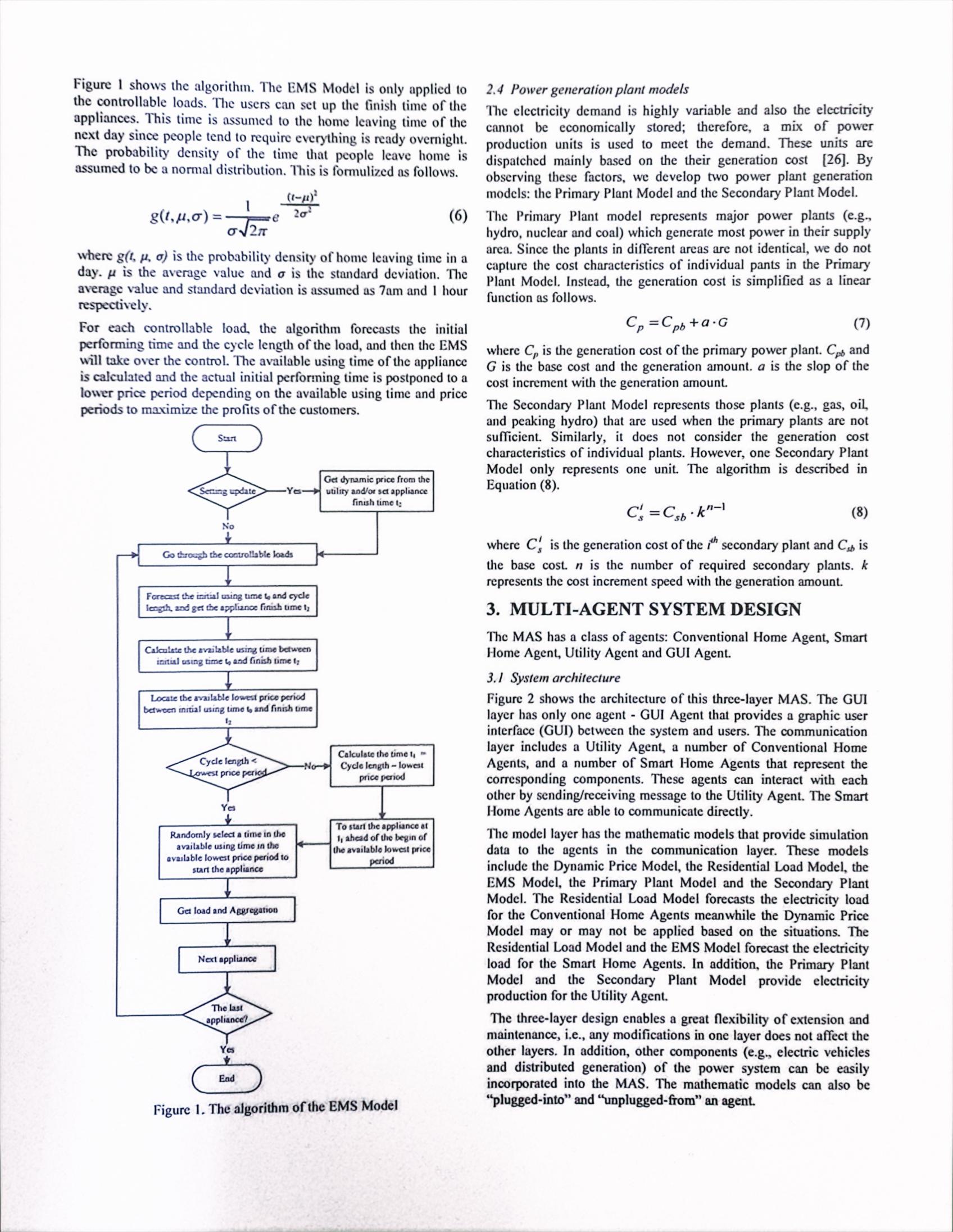}}~
	\subfloat[U-Net]{\includegraphics[width=0.2\textwidth]{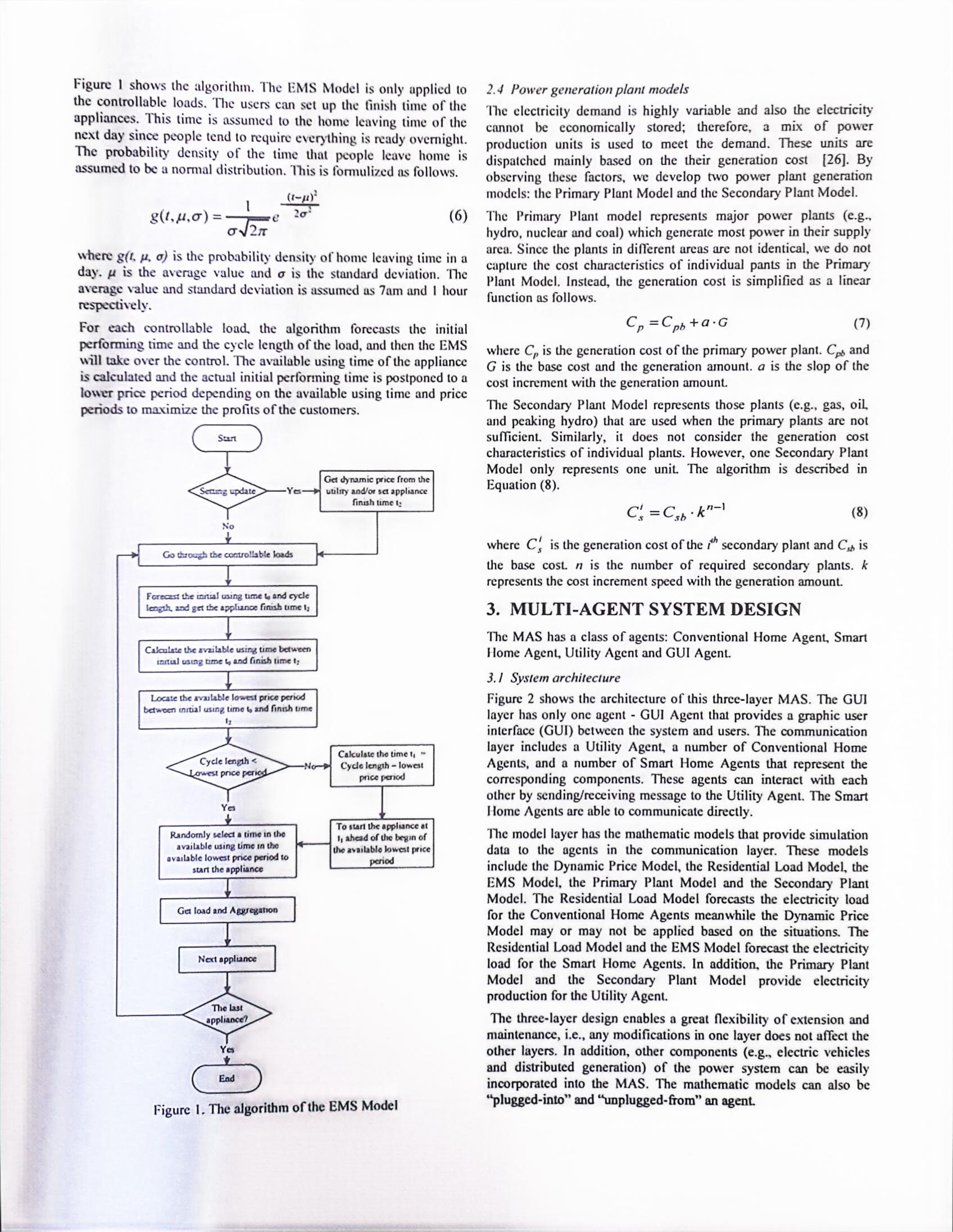}}
	\caption{Results for a black-and-white document image captured under poor lighting conditions}
	\label{fig:good_4}
\end{figure}

The remaining networks that were tested (U-Net~\citep{unet}, ESDR~\citep{edsr}, WDSR-A~\citep{wdsr} and SR-GAN~\citep{srgan}) produced better results in general. Figures~\ref{fig:good_1},~\ref{fig:good_2},~\ref{fig:good_3},~and~\ref{fig:good_4} show a visual comparison of some of their results with the outputs from the algorithm from Fan~\citep{pagelift} as well. The IQA values for those networks and Fan's~\citep{pagelift} (used as a baseline for comparison) in our test set are presented in Table~\ref{tab:result_IQA}. The networks achieved similar results to those of \cite{pagelift}. In some cases, the results surpassed the original algorithm in the PSNR and PIE~\citep{pie} metrics, even though this is the same algorithm used to produce the ground truth images for training the models. These results appear to indicate that the networks successfully generalized the sophisticated operations used in the handcrafted enhancement solution from \cite{pagelift}.

\begin{table}[h]
	\centering
	\caption{Model evaluation using different IQA metrics. The best results are highlighted in bold}
	\begin{tabular}{ccccc}
		\toprule
		\textbf{Engine/Model} & \textbf{PSNR $\uparrow$} & \textbf{MS-SSIM $\uparrow$} & \textbf{WaDIQaM $\uparrow$} & \textbf{PIE $\downarrow$}\\
		\midrule
		\textbf{\cite{pagelift}} & 62.11 & 0.80 & \textbf{0.61} & 0.95\\
		\textbf{U-Net}    & \textbf{63.03} & 0.80 & 0.59 & \textbf{0.89}\\
		\textbf{EDSR}     & 62.56 & 0.80 & 0.57 & 0.95\\
		\textbf{WDSR A}   & 62.56 & 0.80 & 0.58 & 1.1\\
		\textbf{SRGAN}    & 62.56 & 0.80 & 0.55 & 1.2\\
		\bottomrule
	\end{tabular}
	\label{tab:result_IQA}
\end{table}

\section{Conclusions} \label{sec:conclusion}
We investigated deep learning-based solutions for enhancing document images using six different neural network architectures. The trained models were evaluated both qualitatively (\emph{i.e.} with visual inspection) and quantitatively (\emph{i.e.} using both traditional and modern IQA metrics). Moreover, we also compared the results with a traditional computer vision-based approach specifically designed for document image enhancement~\cite{pagelift}.

Based on the results presented in Section~\ref{sec:results}, it is noticeable that deep learning techniques are capable of producing similar results compared to image processing and computer vision based solutions. We believe these are promising results, because they show that deep neural networks are capable of learning powerful representations of document images, which typically have unusual patterns compared to the more common use case of natural images. Furthermore, results from super resolution-based techniques have provided important insights for designing more robust feature extractors for document enhancement. In summary, our results hint that deep learning-based approaches can be a promising alternative for document enhancement (as already showed by~\cite{deepotsu, doc_enhancement}), and can be trained using non-ideal data (as is the case in this paper, where we used pairs of raw and enhanced images provided by~\cite{pagelift} for training). Moreover, the best performing architectures can also be modified to create more robust versions of it for the specific document enhancement task.

A next natural step for future investigation is to fine-tune the hyper-parameters of the neural networks, and look for better ways to leverage their strengths in order to build a robust end-to-end model for document enhancement. To further improve the quality of the models, we also intend to increase the size of the PDF dataset described in Section~\ref{sec:experimental-setup} by automating part of the mechanical process of printing and re-scanning digital documents.

\section*{Acknowledgments}
This paper was achieved in cooperation with HP Inc.\ R\&D Brazil, using incentives of the Brazilian Informatics Law (Law n\textdegree{}.~8.2.48 of 1991). We would like to thank Erasmo Isotton for the management support during production of this work, and Rafael Borges, Sebastien Tandel, Thomas Paula and Juliano Vacaro for their help and discussions on the subject of this paper.

\bibliographystyle{unsrtnat}
\bibliography{template}  

\begin{thebibliography}{41}
\providecommand{\natexlab}[1]{#1}
\providecommand{\url}[1]{\texttt{#1}}
\expandafter\ifx\csname urlstyle\endcsname\relax
  \providecommand{\doi}[1]{doi: #1}\else
  \providecommand{\doi}{doi: \begingroup \urlstyle{rm}\Url}\fi

\bibitem[Pang et~al.(2021)Pang, Lin, Qin, and Chen]{image_translation}
Yingxue Pang, Jianxin Lin, Tao Qin, and Zhibo Chen.
\newblock Image-to-image translation: Methods and applications.
\newblock \emph{arXiv preprint arXiv:2101.08629}, 2021.

\bibitem[Tian et~al.(2020)Tian, Fei, Zheng, Xu, Zuo, and Lin]{img_denoiser}
Chunwei Tian, Lunke Fei, Wenxian Zheng, Yong Xu, Wangmeng Zuo, and Chia-Wen
  Lin.
\newblock Deep learning on image denoising: An overview.
\newblock \emph{Neural Networks}, 2020.

\bibitem[Cheng and Shi(2004)]{image_enhancement}
HD~Cheng and XJ~Shi.
\newblock A simple and effective histogram equalization approach to image
  enhancement.
\newblock \emph{Digital signal processing}, 14\penalty0 (2):\penalty0 158--170,
  2004.

\bibitem[Ronneberger et~al.(2015)Ronneberger, Fischer, and Brox]{unet}
Olaf Ronneberger, Philipp Fischer, and Thomas Brox.
\newblock U-net: Convolutional networks for biomedical image segmentation.
\newblock In \emph{International Conference on Medical image computing and
  computer-assisted intervention}, pages 234--241. Springer, 2015.

\bibitem[Lim et~al.(2017)Lim, Son, Kim, Nah, and Mu~Lee]{edsr}
Bee Lim, Sanghyun Son, Heewon Kim, Seungjun Nah, and Kyoung Mu~Lee.
\newblock Enhanced deep residual networks for single image super-resolution.
\newblock In \emph{Proceedings of the IEEE conference on computer vision and
  pattern recognition workshops}, pages 136--144, 2017.

\bibitem[Yu et~al.(2018)Yu, Fan, Yang, Xu, Wang, Wang, and Huang]{wdsr}
Jiahui Yu, Yuchen Fan, Jianchao Yang, Ning Xu, Zhaowen Wang, Xinchao Wang, and
  Thomas Huang.
\newblock Wide activation for efficient and accurate image super-resolution.
\newblock \emph{arXiv preprint arXiv:1808.08718}, 2018.

\bibitem[Ledig et~al.(2017)Ledig, Theis, Husz{\'a}r, Caballero, Cunningham,
  Acosta, Aitken, Tejani, Totz, Wang, et~al.]{srgan}
Christian Ledig, Lucas Theis, Ferenc Husz{\'a}r, Jose Caballero, Andrew
  Cunningham, Alejandro Acosta, Andrew Aitken, Alykhan Tejani, Johannes Totz,
  Zehan Wang, et~al.
\newblock Photo-realistic single image super-resolution using a generative
  adversarial network.
\newblock In \emph{Proceedings of the IEEE conference on computer vision and
  pattern recognition}, pages 4681--4690, 2017.

\bibitem[Zhang et~al.(2017)Zhang, Zuo, Gu, and Zhang]{ircnn}
Kai Zhang, Wangmeng Zuo, Shuhang Gu, and Lei Zhang.
\newblock Learning deep cnn denoiser prior for image restoration.
\newblock In \emph{Proceedings of the IEEE conference on computer vision and
  pattern recognition}, pages 3929--3938, 2017.

\bibitem[Zhu et~al.(2017)Zhu, Park, Isola, and Efros]{cyclegan}
Jun-Yan Zhu, Taesung Park, Phillip Isola, and Alexei~A Efros.
\newblock Unpaired image-to-image translation using cycle-consistent
  adversarial networks.
\newblock In \emph{Proceedings of the IEEE international conference on computer
  vision}, pages 2223--2232, 2017.

\bibitem[Fan(2007)]{pagelift}
Jian Fan.
\newblock Enhancement of camera-captured document images with watershed
  segmentation.
\newblock \emph{CBDAR07}, pages 87--93, 2007.

\bibitem[Wang et~al.(2003)Wang, Simoncelli, and Bovik]{ssim}
Zhou Wang, Eero~P Simoncelli, and Alan~C Bovik.
\newblock Multiscale structural similarity for image quality assessment.
\newblock In \emph{The Thrity-Seventh Asilomar Conference on Signals, Systems
  \& Computers, 2003}, volume~2, pages 1398--1402. Ieee, 2003.

\bibitem[Prashnani et~al.(2018)Prashnani, Cai, Mostofi, and Sen]{pie}
Ekta Prashnani, Hong Cai, Yasamin Mostofi, and Pradeep Sen.
\newblock Pieapp: Perceptual image-error assessment through pairwise
  preference.
\newblock In \emph{The IEEE Conference on Computer Vision and Pattern
  Recognition (CVPR)}, June 2018.

\bibitem[Bosse et~al.(2017)Bosse, Maniry, M{\"u}ller, Wiegand, and
  Samek]{wadiqam}
Sebastian Bosse, Dominique Maniry, Klaus-Robert M{\"u}ller, Thomas Wiegand, and
  Wojciech Samek.
\newblock Deep neural networks for no-reference and full-reference image
  quality assessment.
\newblock \emph{IEEE Transactions on Image Processing}, 27\penalty0
  (1):\penalty0 206--219, 2017.

\bibitem[Mohammadi et~al.(2014)Mohammadi, Ebrahimi-Moghadam, and Shirani]{iqa}
Pedram Mohammadi, Abbas Ebrahimi-Moghadam, and Shahram Shirani.
\newblock Subjective and objective quality assessment of image: A survey.
\newblock \emph{arXiv preprint arXiv:1406.7799}, 2014.

\bibitem[Niu et~al.(2018)Niu, Zhong, Guo, Shi, and Chen]{iqa_survey}
Yuzhen Niu, Yini Zhong, Wenzhong Guo, Yiqing Shi, and Peikun Chen.
\newblock 2d and 3d image quality assessment: A survey of metrics and
  challenges.
\newblock \emph{IEEE Access}, 7:\penalty0 782--801, 2018.

\bibitem[Priyadarshini et~al.(2021)Priyadarshini, Bharani, Rahimankhan, and
  Rajendran]{img_enh1}
R~Priyadarshini, Arvind Bharani, E~Rahimankhan, and N~Rajendran.
\newblock Low-light image enhancement using deep convolutional network.
\newblock In \emph{Innovative Data Communication Technologies and Application},
  pages 695--705. Springer, 2021.

\bibitem[Tao et~al.(2018)Tao, Yang, Wu, Liu, Zhou, and Liu]{img_enh2}
Fuyu Tao, Xiaomin Yang, Wei Wu, Kai Liu, Zhili Zhou, and Yiguang Liu.
\newblock Retinex-based image enhancement framework by using region covariance
  filter.
\newblock \emph{Soft Computing}, 22\penalty0 (5):\penalty0 1399--1420, 2018.

\bibitem[Vu et~al.(2018)Vu, Van~Nguyen, Pham, Luu, and Yoo]{img_enh3}
Thang Vu, Cao Van~Nguyen, Trung~X Pham, Tung~M Luu, and Chang~D Yoo.
\newblock Fast and efficient image quality enhancement via desubpixel
  convolutional neural networks.
\newblock In \emph{Proceedings of the European Conference on Computer Vision
  (ECCV) Workshops}, pages 0--0, 2018.

\bibitem[Wang et~al.(2020)Wang, Chen, and Hoi]{superresolution_survey}
Zhihao Wang, Jian Chen, and Steven~CH Hoi.
\newblock Deep learning for image super-resolution: A survey.
\newblock \emph{IEEE transactions on pattern analysis and machine
  intelligence}, 2020.

\bibitem[Baek(2016)]{dropbox}
Jongmin Baek.
\newblock Fast document rectification and enhancement.
\newblock Available at:
  https://dropbox.tech/machine-learning/fast-document-rectification-and-enhancement,
  2016.
\newblock Accessed in: 2020-05-13.

\bibitem[He and Schomaker(2019)]{deepotsu}
Sheng He and Lambert Schomaker.
\newblock Deepotsu: Document enhancement and binarization using iterative deep
  learning.
\newblock \emph{Pattern Recognition}, 91:\penalty0 379--390, 2019.

\bibitem[Hidalgo et~al.(2005)Hidalgo, Espana, Castro, and
  P{\'e}rez]{doc_enhancement}
Jos{\'e}~Luis Hidalgo, Salvador Espana, Mar{\'\i}a~Jos{\'e} Castro, and
  Jos{\'e}~Alberto P{\'e}rez.
\newblock Enhancement and cleaning of handwritten data by using neural
  networks.
\newblock In \emph{Iberian Conference on Pattern Recognition and Image
  Analysis}, pages 376--383. Springer, 2005.

\bibitem[Gao et~al.(2010)Gao, Lu, Tao, and Li]{iqa_hvs}
Xinbo Gao, Wen Lu, Dacheng Tao, and Xuelong Li.
\newblock Image quality assessment and human visual system.
\newblock In \emph{Visual Communications and Image Processing 2010}, volume
  7744, page 77440Z. International Society for Optics and Photonics, 2010.

\bibitem[Li et~al.(2019)Li, Zhu, and Qiu]{doc_iqa}
Hongyu Li, Fan Zhu, and Junhua Qiu.
\newblock Towards document image quality assessment: A text line based
  framework and a synthetic text line image dataset.
\newblock \emph{arXiv preprint arXiv:1906.01907}, 2019.

\bibitem[Shen et~al.(2019)Shen, Salehi, Baldwin, and Qi]{doc_iqa2}
Aili Shen, Bahar Salehi, Timothy Baldwin, and Jianzhong Qi.
\newblock A joint model for multimodal document quality assessment.
\newblock In \emph{2019 ACM/IEEE Joint Conference on Digital Libraries (JCDL)},
  pages 107--110. IEEE, 2019.

\bibitem[Hussain et~al.(2018)Hussain, Wahab, Idris, Ho, and Jung]{diqa1}
Mehdi Hussain, Ainuddin Wahid~Abdul Wahab, Yamani Idna~Bin Idris, Anthony~TS
  Ho, and Ki-Hyun Jung.
\newblock Image steganography in spatial domain: A survey.
\newblock \emph{Signal Processing: Image Communication}, 65:\penalty0 46--66,
  2018.

\bibitem[Alaei et~al.(2018)Alaei, Raveaux, Conte, and Stantic]{diqa2}
Alireza Alaei, Romain Raveaux, Donatello Conte, and Bela Stantic.
\newblock " quality" vs." readability" in document images: Statistical analysis
  of human perception.
\newblock In \emph{2018 13th IAPR International Workshop on Document Analysis
  Systems (DAS)}, pages 363--368. IEEE, 2018.

\bibitem[Chollet(2015)]{keras}
Francois et~al. Chollet.
\newblock Keras, 2015.
\newblock URL \url{https://github.com/fchollet/keras}.

\bibitem[et~al.(2015)]{tensorflow}
Mart\'{\i}n~Abadi et~al.
\newblock {TensorFlow}: Large-scale machine learning on heterogeneous systems,
  2015.
\newblock URL \url{http://tensorflow.org/}.
\newblock Software available from tensorflow.org.

\bibitem[Ma et~al.(2018)Ma, Shu, Bai, Wang, and Samaras]{docunet}
Ke~Ma, Zhixin Shu, Xue Bai, Jue Wang, and Dimitris Samaras.
\newblock Docunet: document image unwarping via a stacked u-net.
\newblock In \emph{Proceedings of the IEEE Conference on Computer Vision and
  Pattern Recognition}, pages 4700--4709, 2018.

\bibitem[Hu et~al.(2019)Hu, Naiel, Wong, Lamm, and Fieguth]{runet}
Xiaodan Hu, Mohamed~A Naiel, Alexander Wong, Mark Lamm, and Paul Fieguth.
\newblock Runet: A robust unet architecture for image super-resolution.
\newblock In \emph{Proceedings of the IEEE/CVF Conference on Computer Vision
  and Pattern Recognition Workshops}, pages 0--0, 2019.

\bibitem[Huang et~al.(2018)Huang, Zhu, Geng, Ran, Zhou, Xing, Wan, and
  Ji]{app_unet}
Jie Huang, Pengfei Zhu, Mingrui Geng, Jiewen Ran, Xingguang Zhou, Chen Xing,
  Pengfei Wan, and Xiangyang Ji.
\newblock Range scaling global u-net for perceptual image enhancement on mobile
  devices.
\newblock In \emph{Proceedings of the European Conference on Computer Vision
  (ECCV) Workshops}, pages 0--0, 2018.

\bibitem[Bank et~al.(2020)Bank, Koenigstein, and Giryes]{autoencoder}
Dor Bank, Noam Koenigstein, and Raja Giryes.
\newblock Autoencoders.
\newblock \emph{arXiv preprint arXiv:2003.05991}, 2020.

\bibitem[Kingma and Ba(2014)]{adam}
Diederik~P Kingma and Jimmy Ba.
\newblock Adam: A method for stochastic optimization.
\newblock \emph{arXiv preprint arXiv:1412.6980}, 2014.

\bibitem[He et~al.(2016)He, Zhang, Ren, and Sun]{resnet}
Kaiming He, Xiangyu Zhang, Shaoqing Ren, and Jian Sun.
\newblock Deep residual learning for image recognition.
\newblock In \emph{Proceedings of the IEEE conference on computer vision and
  pattern recognition}, pages 770--778, 2016.

\bibitem[Salimans and Kingma(2016{\natexlab{a}})]{weights_normalization}
Tim Salimans and Durk~P Kingma.
\newblock Weight normalization: A simple reparameterization to accelerate
  training of deep neural networks.
\newblock In \emph{Advances in neural information processing systems}, pages
  901--909, 2016{\natexlab{a}}.

\bibitem[Salimans and Kingma(2016{\natexlab{b}})]{weightnorm}
Tim Salimans and Diederik~P. Kingma.
\newblock Weight normalization: A simple reparameterization to accelerate
  training of deep neural networks.
\newblock In \emph{Neural Information Processing Systems 2016},
  2016{\natexlab{b}}.

\bibitem[Qassim et~al.(2018)Qassim, Verma, and Feinzimer]{vgg}
Hussam Qassim, Abhishek Verma, and David Feinzimer.
\newblock Compressed residual-vgg16 cnn model for big data places image
  recognition.
\newblock In \emph{2018 IEEE 8th Annual Computing and Communication Workshop
  and Conference (CCWC)}, pages 169--175. IEEE, 2018.

\bibitem[Deng et~al.(2009)Deng, Dong, Socher, Li, Li, and Fei-Fei]{imagenet}
Jia Deng, Wei Dong, Richard Socher, Li-Jia Li, Kai Li, and Li~Fei-Fei.
\newblock Imagenet: A large-scale hierarchical image database.
\newblock In \emph{2009 IEEE conference on computer vision and pattern
  recognition}, pages 248--255. Ieee, 2009.

\bibitem[Goodfellow et~al.(2014)Goodfellow, Pouget-Abadie, Mirza, Xu,
  Warde-Farley, Ozair, Courville, and Bengio]{gans}
Ian Goodfellow, Jean Pouget-Abadie, Mehdi Mirza, Bing Xu, David Warde-Farley,
  Sherjil Ozair, Aaron Courville, and Yoshua Bengio.
\newblock Generative adversarial nets.
\newblock In \emph{Advances in neural information processing systems}, pages
  2672--2680, 2014.

\bibitem[Howard et~al.(2017)Howard, Zhu, Chen, Kalenichenko, Wang, Weyand,
  Andreetto, and Adam]{mobilenetv1}
Andrew~G Howard, Menglong Zhu, Bo~Chen, Dmitry Kalenichenko, Weijun Wang,
  Tobias Weyand, Marco Andreetto, and Hartwig Adam.
\newblock Mobilenets: Efficient convolutional neural networks for mobile vision
  applications.
\newblock \emph{arXiv preprint arXiv:1704.04861}, 2017.

\end{thebibliography}

~
~
~

\appendix

\section{IQA Investigation} \label{sec:iqa_investigation}

In order to evaluate which full-reference IQA metrics would be better suited for evaluating document image quality, we first used the algorithm from \cite{pagelift} to enhance the captured images from our test set and manually selected the images that performed best using the same methodology presented in Section~\ref{sec:iqa_methodology}. We then compared each raw, enhanced and white images (used as a control image, as the dataset comprises only documents with white background) with their corresponding ground truth (the digital PDF version), storing the values returned by each of the metrics.

For this work, we assume that a proper quality metric should return better scores (\emph{i.e.} return higher or lower value, depending on the metric) on enhanced images than on their captured (raw) or fully white counterparts. Similarly, the captured image should have better scores than the fully white image (as the white image can be seen as the original with all of its content erased). Therefore, for each metric, we defined two error measures:
\begin{description}
    \item[raw errors] as how many times the captured image produced a better score than the enhanced image;
    \item[white errors] as how many times the fully white image produced a better score than the captured image.
\end{description}

The results are presented in Table~\ref{tab:iqa_compare}\footnote{In the PDF version of this paper, you can click on each metric to navigate to the implementation used herein.}. The metrics with the lowest total error (sum of \emph{raw} and \emph{white} errors) were, respectively: PIE~\cite{pie}, WaDIQaM~\cite{wadiqam} and the MS-SSIM~\cite{ssim}. For this reason, we used these metrics as a baseline to compare our results. We also used the PSNR error, because it is a traditional IQA metric, and is based on the MSE error that was used to train most of the neural networks studied in this work.

\begin{table}[htb]
    \centering
	\caption{Relative values from IQA results. Lower values indicate better performance. The best results are highlighted in bold. The \emph{Total} column is the sum of the \emph{Raw error} and the \emph{White error} for each row.}
	\begin{tabular}{cccc}
		\toprule
		\textbf{Metric} & \textbf{Raw error} & \textbf{White error} & \textbf{Total}\\
		\midrule
		\href{https://github.com/bwohlberg/sporco/blob/master/sporco/metric.py}{BSNR} & 4.0 & 16.2 & 20.2\\
		\href{https://scikit-image.org/docs/dev/api/skimage.measure.html#skimage.measure.compare_psnr}{PSNR} & 0 & 28.8 & 28.8 \\
		\href{https://github.com/bwohlberg/sporco/blob/master/sporco/metric.py}{MAE} & \textbf{0} & 38.4 & 38.4\\
		\href{https://scikit-image.org/docs/dev/api/skimage.measure.html#skimage.measure.compare_nrmse}{MSE} & \textbf{0} & 28.8 & 28.8\\
		\href{https://github.com/bwohlberg/sporco/blob/master/sporco/metric.py}{SNR} & \textbf{0} & 28.8 & 28.8\\
		\href{https://github.com/bwohlberg/sporco/blob/master/sporco/metric.py}{PAMSE} & \textbf{0} & 10.6 & 10.6\\
		\href{https://scikit-image.org/docs/dev/api/skimage.measure.html#skimage.measure.compare_ssim}{MS-SSIM} & 2.5 & 1.0 & 3.5\\
		\href{https://sewar.readthedocs.io/en/latest/}{ERGAS} & 6.1 & 22.2 & 28.3 \\
		\href{https://sewar.readthedocs.io/en/latest/}{RMSE} & \textbf{0} & 33.3 & 33.3\\
		\href{https://sewar.readthedocs.io/en/latest/}{RMSE-SW} & 0.5 & 36.9 & 41.9\\
		\href{https://sewar.readthedocs.io/en/latest/}{SAM} & 35.9 & 37.4 & 73.3\\
		\href{https://sewar.readthedocs.io/en/latest/}{SCC} & 22.7 & \textbf{0} & 22.7\\
		\href{https://sewar.readthedocs.io/en/latest/}{UQI} & 0.5 & 37.4 & 42.4\\
		\href{https://sewar.readthedocs.io/en/latest/}{VIFP} & 7.1 & \textbf{0} & 7.1\\
		\href{http://imageprocessing-sankarsrin.blogspot.com/2018/06/comprehensive-survey-on-full-reference.html}{MAD} & 49.0 & 0.5 & 49.5\\
		\href{http://imageprocessing-sankarsrin.blogspot.com/2018/06/comprehensive-survey-on-full-reference.html}{F-SSIM} & 15.2 & \textbf{0} & 15.2\\
		\href{http://imageprocessing-sankarsrin.blogspot.com/2018/06/comprehensive-survey-on-full-reference.html}{PSNR HMA} & 6.6 & \textbf{0} & 6.6\\
		\href{http://imageprocessing-sankarsrin.blogspot.com/2018/06/comprehensive-survey-on-full-reference.html}{SR SIM} & 37.4 & \textbf{0} & 37.4\\
		\href{http://imageprocessing-sankarsrin.blogspot.com/2018/06/comprehensive-survey-on-full-reference.html}{GMSD} & 30.8 & 100 & 130.8 \\
		\href{http://imageprocessing-sankarsrin.blogspot.com/2018/06/comprehensive-survey-on-full-reference.html}{MDSI} & 6.1 & \textbf{0} & 6.1\\
		\href{http://imageprocessing-sankarsrin.blogspot.com/2018/06/comprehensive-survey-on-full-reference.html}{HaarPSI} & 35.4 & \textbf{0} & 35.4\\
		\href{https://github.com/prashnani/PerceptualImageError}{PIE} & 0.5 & \textbf{0} & \textbf{0.5}\\
		\href{https://github.com/lidq92/WaDIQaM}{WaDIQaM} & \textbf{0} & 3.0 & 3.0\\
		\bottomrule
	\end{tabular}
	\label{tab:iqa_compare}
\end{table}

\end{document}